\documentclass[12pt]{article}

\usepackage{amsmath,amssymb}
\usepackage{subfiles} 
\usepackage{bm} 
\usepackage{upgreek}
\usepackage{todonotes}
\usepackage{multirow}
\usepackage{multicol}
 \usepackage{booktabs}
\usepackage{bbm, dsfont}

\newcommand{\mf}{\mathbf}

\newcommand{\N}{\mathcal{N}}

\newcommand{\mxr}{MixEHR-S}

 \usepackage{authblk}
  \usepackage{url}

\begin{document}

\title{Supervised multi-specialist topic model with applications on large-scale electronic health record data}


\author[1]{Ziyang Song}
\author[1]{Xavier Sumba Toral}
\author[1]{Yixin Xu}
\author[2]{Aihua Liu}
\author[2]{Liming Guo}
\author[3]{Guido Powell}
\author[3]{Aman Verma}
\author[3,*]{David Buckeridge}
\author[2,*]{Ariane Marelli}
\author[1,*]{Yue Li}
\affil[1]{School of Computer Science, McGill University, Montreal, QC, Canada, H3A 2A7}
\affil[2]{McGill Adult Unit for Congenital Heart Disease Excellence, Montreal, QC, Canada, H4A 3J1}
\affil[3]{School of Population and Global Health, McGill University, Montreal, QC, Canada, H3A1A3}
\affil[*]{Correspondence to david.buckeridge@mcgill.ca, ariane.marelli@mcgill.ca, yueli@cs.mcgill.ca}

\maketitle

\begin{abstract}
\textbf{Motivation:} Electronic health record (EHR) data provides a new venue to elucidate disease comorbidities and latent phenotypes for precision medicine. To fully exploit its potential, a realistic data generative process of the EHR data needs to be modelled.\\
\textbf{Materials and Methods:} We present \mxr~to jointly infer specialist-disease topics from the EHR data. As the key contribution, we model the specialist assignments and ICD-coded diagnoses as the latent topics based on patient's underlying disease topic mixture in a novel unified supervised hierarchical Bayesian topic model. For efficient inference, we developed a closed-form collapsed variational inference algorithm to learn the model distributions of \mxr. \\
\textbf{Results:} We applied \mxr~to two independent large-scale EHR databases in Quebec with three targeted applications: (1) Congenital Heart Disease (CHD) diagnostic prediction among 154,775 patients; (2) Chronic obstructive pulmonary disease (COPD) diagnostic prediction among 73,791 patients; (3) future insulin treatment prediction among 78,712 patients diagnosed with diabetes as a mean to assess the disease exacerbation. In all three applications, \mxr~conferred clinically meaningful latent topics among the most predictive latent topics and achieved superior target  prediction accuracy compared to the existing methods, providing opportunities for prioritizing high-risk patients for healthcare services.\\
\textbf{Availability and implementation:} \mxr~source code and scripts of the experiments are freely available at \url{https://github.com/li-lab-mcgill/mixehrS}

\end{abstract}

\section{Introduction}
With the rapid adoption of electronic health record (EHR), there is an unprecedented opportunity to re-define medical concepts and automate disease diagnosis process. EHR include standardized digital codes such as the International Classification of Diseases (ICD) 9 codes, which often span over tens of thousands of features. Traditional statistical methods are incapable of handling the high-dimensional EHR features and therefore often require engineering a small set of hand-crafted features. Topic models, on the other hand, show great promise in representing the entire discrete EHR data by a set of latent topics \cite{blei2003latent,Chen2015-yx,Pivovarov2015-xf,Li:2020exa}. In analogy to text categorization \cite{blei2003latent}, we consider the EHR history for each patient as a document, which exhibits a mixture of memberships over a set of latent disease topics. However, existing EHR topic modeling ignores the generative process of the clinical specialists in diagnosing the patient. Also, most of EHR topic models are unsupervised and therefore require a downstream supervised classifier to perform prediction of a target disease. 

In this paper, we present \mxr~as a novel unified supervised multi-specialist Bayesian topic model. \mxr~stands out from the existing methods with three key contributions. First, we explicitly model the distribution of specialists based on the patient's latent disease topic mixture. Second, we infer the specialist-specific latent disease topics, which capture the different clinical domain knowledge. Third, to predict a binary target labels such as a disease diagnosis, we developed a Bayesian probit regression component to form a supervised topic model. This allows us to learn a linear classifier to predict a binary label using the topic mixture inferred for each patient. The posterior inference of the latent disease topics for each patient's diagnosis in turn takes into account the predictive likelihood of the target disease. Therefore, the topic inference step and the supervised learning step can benefit from each other during the model training. 

Using real-world large-scale EHR databases in Quebec, we demonstrated the utility of \mxr~model with three targeted applications: (1) Congenital Heart Disease (CHD) diagnostic prediction among 154,775 patients; (2) Chronic obstructive pulmonary disease (COPD) diagnostic prediction among 73,791 patients; (3) future insulin treatment prediction among 78,712 patients diagnosed with diabetes. In all three applications, we observe clinically meaningful latent topics among the most predictive latent topics and achieved superior target  prediction accuracy compared to the existing methods, providing opportunities for prioritizing high-risk patients and drug recommendations.

\section{Related Methods}
Our \mxr~model is related to the best-known topic model Latent Dirichlet Allocation (LDA) \cite{blei2003latent}, which has been applied to raw clinical text for medical tasks in the past \cite{Chen2015-yx}. However, LDA is often inadequate to model complex diagnoses because it does not account for heterogeneous data categories. The mutli-view topic model UPhenome could learn diseases and patient characteristics with a fixed set of data types for heterogeneous medical records \cite{Pivovarov2015-xf}. However, UPhenome is unable to infer specialist-diagnosis mechanisms as presented in our \mxr.

Our Bayesian approach is also related to the frequentest non-negative matrix factorization (NMF) methods that were seen applications in the EHR domain. In particular, Joshi et al. (2016) described a NMF model to identify multiple co-occurring medical conditions using clinical notes  \cite{Joshi:2016ty}. Extending the single-view NMF framework, Gunasekar et al. (2016) proposed a collective NMF model called SiCNMF for modeling multi–source EHR Data \cite{Gunasekar2016-gr}. Compared to the Bayesian topic modeling, these models are often less interpretable because they do not model the data generative processes.

In our recent work, we described a multi-modal topic model called MixEHR \cite{Li:2020exa}, which models different EHR data types with distinct categorical distributions. MixEHR is an unsupervised topic model, which needs training an additional classifier to predict a specific target label. One solution to this problem is to use supervised latent Dirichlet allocation (sLDA) \cite{NIPS2007_3328}, which infers latent topic distributions over documents (e.g., patients) while training a linear regression model to predict a continuous response of the documents (whereas we predict binary outcome with a Probit link). Along this direction, Zhang et al. (2017) proposes survival topic model, a supervised topic model that models jointly patients’ discharge summaries and a Cox hazard ratios on patient mortality with a limited scope \cite{zhang2017survival}. 

Although related, compared to the aforementioned work, our approach departs further from the  recently popularized neural networks including our own \cite{Lu:2021cp} on supervised classification of target clinical outcomes \cite{Razavian2015-qd,Cheng2016-vv,Lipton2015-vg,Choi2016-gl,Nguyen2016-fz,Rajkomar:2018wj}. In these models, prediction accuracy is prioritized over model interpretability. The latter has been a challenge with the distributed representation of the neural networks. It often requires more sophisticated and computationally expensive techniques such as knowledge graph embedding with the attention mechanisms as demonstrated in GRAMS \cite{Choi:2017fma} to improve interpretability and mitigate over-fitting. Additionally, autoencoders \cite{Miotto2016-hn,Suresh2017-xx} such as Deep Patients \cite{Miotto2016-hn} were also applied to learn low-dimensional manifold of the EHR data in an unsupervised fashion. These models often require careful fine-tuning of a large number of network hyperparameters and have similar interpretation challenges as the supervised neural networks.

\section{Methods}
\subsection{\mxr~model generative process}
\begin{figure*}[t]
  \centering\includegraphics[width=\linewidth]{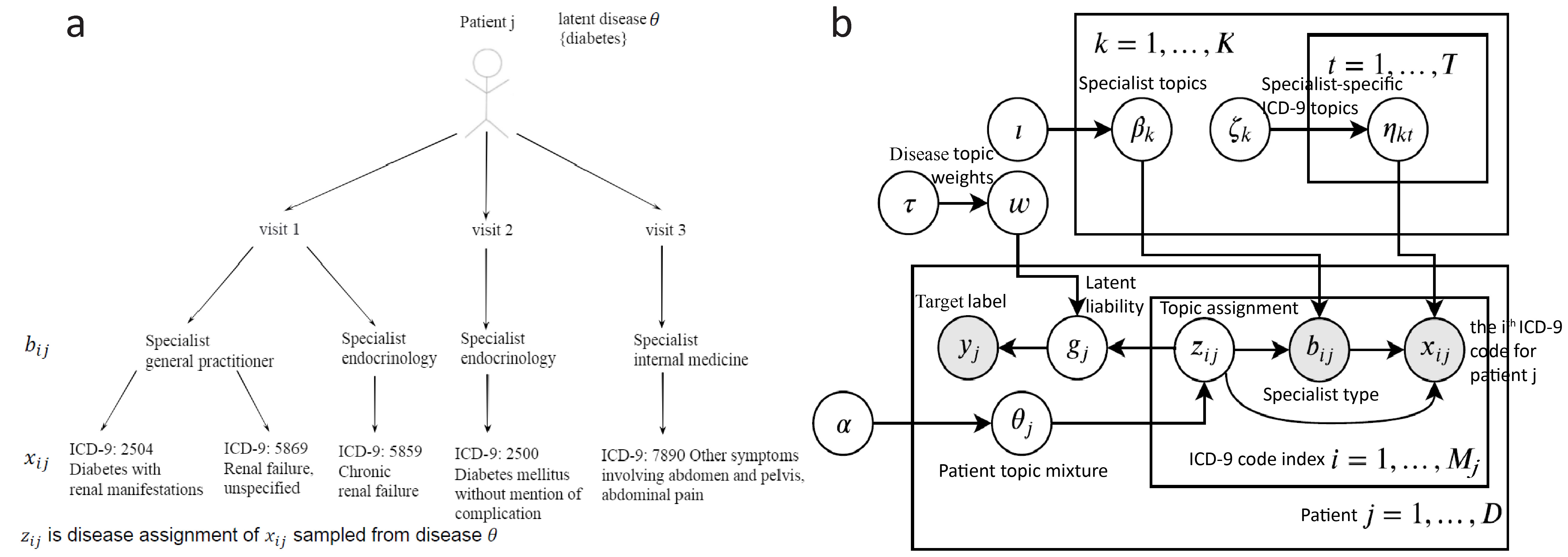}
  \caption{\mxr~model overview.
  (a) Conceptual illustration of the multi-specialist EHR data generative process. A diabetes patient $j$ made three outpatient visits to receive a total of five ICD-9 diagnosis codes $i\in\{1,\ldots,5\}$. Each code has its own underlying disease cause or topic $z_{ij}$, which was sampled from the patient disease mixture $\theta_j$. Based on the disease cause, a specialist $b_{ij}$ was  assigned to the patient. The actual diagnosis code $x_{ij}$ was made by the specialist based on his expertise.
  (b) The plate model for the multi-specialist EHR model. The probabilistic graphical model (PGM) formally characterizes the data generative process. The observed variables are the shaded nodes, including the target label $y_j$, specialist type $b_{ij}$ and diagnosis code $x_{ij}$. The unshaded nodes are the latent variables.  The details of the variables in the graphical model are described in the main text and summarized in Table.~\ref{table: vardescription}.}
  \label{fig:mixehr_model}
\end{figure*}

\noindent We model the heterogeneous medical data using a generative topic model illustrated in Fig.~\ref{fig:mixehr_model}. For the notation below, we use boldface to denote the vectors, capital letters for constants, and regular case for scalar variables. For each patient $j \in (1, \dots, D)$, we index his/her ICD-9 code by $i \in \{1, \dots, M_j\}$ for $M_j$ total number of ICD diagnose codes. Each ICD-9 code $x_{ij}$ is assigned by one of the specialists $b_{ij}$ with index $t \in \{1, \dots, T \}$. Each patient is also associated with a binary target disease label $y_j$.

We assume that each patient $j$ follows a disease topic mixture $\bm{\theta}_j$, which is a $K$-dimensional Dirichlet distribution $\text{Dir}(\bm{\alpha})$ with unknown hyperparameter $\bm{\alpha}$. To generate an ICD-9 code for a patient, we first sample a latent topic $z_{ij}=k$ from categorical distribution with rate set to $\bm{\theta}_j$. For vectorized notations, we represent the discrete topic assignment $z_{ij}=k$ by a binary one-hot vector $\mf{z}_{ij}$ such that $z_{ijk}=1$ if $z_{ij}=k$ and $z_{ijk'}=0 \forall k'\ne k$.

We then sample a specialist $b_{ij}=t$ from a topic distribution $\bm{\beta}_k \sim \text{Dir}(\iota)$, which is a $T$-dimensional Dirichlet variable with hyperparameter $\iota$ for $T$ specialists. Given the topic assignment $k$ and the specialist assignment $t$, we then sample the ICD-9 code from a categorical distribution with the rate $\bm{\eta}_{kt}$ that follows a set of $V$-dimensional Dirichlet distribution $Dir(\bm{\zeta})$. The Dirichlet hyperparameters $\bm{\alpha}$, $\bm{\iota}$ and $\bm{\zeta}$ are given Gamma priors with fixed parameters. 

To sample the target disease label $y_j$ for patient $j$, we first sample a Gaussian liability variable $g_j \sim \N(\mf{w}^\top \bar{\mf{z}}_j, 1)$, where $\mf{w}$ denotes the global regression coefficient and $\bar{\mf{z}}_j$ denotes the $K$-dimensional topic assignment average over the $M_j$ codes. We then set the target disease label $y_j$ to 1 if $g_j$ is positive otherwise $y_j=0$. This is the Probit regression component of \mxr~model, which can be viewed as a supervised topic model.

Formally, our data generative process starts by generating the global topic distributions over the specialists and ICD-9 codes per specialist for each topic as well as the topic-specific coefficients for the response variable, respectively: 
\begin{equation}\label{eq:global_variables}
    \bm{\beta}_k \sim \text{Dir}(\iota){,\;}
    \bm{\eta}_{kt} \sim \text{Dir}(\zeta_k) {,\;}
    \mf{w} \sim \mathcal{N}(0, \tau^{-1}\mf{I})
\end{equation}

We then sample the local variables, namely the topic mixtures, topic assignments, specialist assignments, and ICD-9 codes for each patient $j$: 
\begin{align} \label{eq:local_variables}
     \bm{\theta}_j &\sim \text{Dir}(\alpha) {,\;}\quad
     z_{ij} \sim \text{Cat}(\bm{\theta}_j) {,\;}\\
     b_{ij} &\sim \text{Cat}(\bm{\beta}_{z_{ij}}) {,\;}\quad
     x_{ij} \sim \text{Cat}(\bm{\eta}_{z_{ij}b_{ij}})
\end{align}

The the hyperparameters are Gamma-distributed:
\begin{equation} \label{eq:hyperparameters}
    \alpha \sim \text{Gamma}(c_{\alpha}, d_{\alpha}) {,\;}
    \iota \sim \text{Gamma}(c_{\iota}, d_{\iota}) {,\;}
    \zeta \sim \text{Gamma}(c_{\zeta}, d_{\zeta})
\end{equation}

The binary label variable is sampled from a Probit distribution:
\begin{align}\label{eq:4}
    g_j \sim \N(\mf{w}^\top\bar{\mf{z}}_j, 1), \quad
    y_j  = \begin{cases} 
        1, & \text{if}\ g_j > 0 \\
        0, & \text{if}\ g_j \leq 0 \end{cases}
\end{align}
where $\bar{\mf{z}}_j = \frac{1}{M_j}\sum^{M_j}_{i=1} \mf{z}_{ij}$ and $\mf{z}_{ij}$ is a $K$-dimensional binary one-hot vector with topic indexed by $z_{ij}$ set to 1 and the rest set to zeros.

\subsection{Model Inference}
Treating the latent variables as missing data, the complete joint likelihood based on the proposed model (Fig.~\ref{fig:mixehr_model}) is $p(\mf{z},\mf{b},\mf{x},\mf{y},\mf{g},\mf{w},\bm{\theta},\bm{\eta},\bm{\beta})$. Exploiting conjugate property of the Dirichlet variables $\bm{\theta}$, $\bm{\beta}$, and $\bm{\eta}$ to the respective categorical variables $\mf{z}$, $\mf{b}$, and $\mf{x}$, we first analytically integrated out the Dirichlet variables given the hyperparameters, resulting in the following marginal joint likelihood:
\begin{align}\label{eq:5}
    & p(\mf{z}, \mf{b},\mf{x},\mf{y}, \mf{g}, \mf{w} \mid \alpha, \iota, \zeta, \tau) 
    \notag\\ 
    = & \int p(\mf{z},\mf{b},\mf{x},\mf{y},\mf{g},\mf{w},\bm{\theta},\bm{\eta},\bm{\beta} \mid \alpha, \iota, \zeta, \tau)d\bm{\theta} d\bm{\beta} d\bm{\eta} \notag\\ 
    =& p(\mf{z},\mf{b},\mf{x} \mid d\alpha, \iota, \zeta) p(\mf{g}|\mf{z},\mf{w})p(\mf{y}|\mf{g})p(\mf{w}|\tau)
\end{align}
The full derivation is described in \textbf{Appendix}~\ref{sec:appendix_full_likelihood}. To approximate the sufficient statistics that are needed for inferring the posterior distribution of the latent variables $p(\mf{z}|\mf{b},\mf{x}, \mf{g})$, we use variational inference \cite{blei2017variational}. In particular, we maximize the Evidence Lower Bound (ELBO):
\begin{align} \label{eq:elbo} 
     \mathcal{L}(\bm{\Theta}) & = 
     \mathbb{E}_{q(\mf{z},\mf{g},\mf{w})} [\log p(\mf{z}, \mf{g},\mf{w}, \mf{b}, \mf{x}, \mf{y})]\notag\\ 
     &- \mathbb{E}_{q(\mf{z}, \mf{g},\mf{w})} [\log q(\mf{z}, \mf{g},\mf{w})]
\end{align}
Under the mean-field factorization, the proposed distribution of the latent variables have the same distributions as their priors: 
\begin{align}
    q(\mf{z}, \mf{g}, \mf{w} \mid \mf{m}, \mf{S}, \bm{\gamma}) = \N(\mf{w} \mid \mf{m}, \mf{S}) 
    \prod_{ijk} \gamma_{ijk}^{[z_{ij}=k]}
    \prod_jq(g_j \mid \lambda_j, 1)
\end{align}
where $\bar{\bm{\gamma}}_j = \frac{1}{M_j}\sum_i\bm{\gamma}_{ij}$ and $\bm{\gamma}_{ij}$ is a $K$-dimensional vector for the $K$ topics. Maximizing ELBO with respect to the variational parameters is equivalent to calculating the expectations \cite{Bishop:2006ui}: $\mathbb{E}_{q(\mf{z}^{-(i,j)},\mf{g})}[\ln q(\mf{z}_{ij} | \bm{\gamma})]$, $\mathbb{E}_{q(\mf{z},\mf{g})}[\ln q(\mf{w} | \mf{m},\mf{S})]$, and $\mathbb{E}_{q(\mf{z},\mf{w})}[\ln q(\mf{g} | \mf{m}, \mf{z})]$. Here $\mf{z}^{-(i,j)}$ denotes all of the latent variables except for variable $\mf{z}_{ij}$. 

For the Bayesian regression component of the \mxr~model, we posit a truncated Gaussian distribution $\mathcal{TN}$ with mean $\bm{\lambda}$ and fixed variance for the liability variable $\mf{g}$:
\begin{equation} \label{eq:var_q_g}
    \begin{aligned}
    & q(g_j \mid \lambda_j) = \begin{cases}
    \mathcal{TN}_+(\lambda_j, 1), & \text{if $y_j=1$}. \\
    \mathcal{TN}_-(\lambda_j, 1), & \text{if $y_j=0$}.
    \end{cases}
    \end{aligned}
\end{equation}
where 
\begin{align}\label{eq:trunc_norm}
    \lambda_j  = \mathbb{E}_{q(\mf{w})}[\mf{w}^\top] \mathbb{E}_{q(\mf{z})}[\bar{\mf{z}}_j]
\end{align}
The variational distribution of the regression coefficients $\mf{w}$ follows a multivariate Gaussian distribution:
\begin{align}\label{eq:var_q_w}
    q(\mf{w} \mid \mf{m}, \mf{S}) = \mathcal{N}(\mf{m}, \mf{S})
\end{align}
where the variational parameters for $q(\mf{w} \mid \mf{m}, \mf{S})$ can be solved by completing the square (\textbf{Appendix}~\ref{sec:appendix_full_likelihood}):
\begin{align} \label{eq:para_update_w}
    \mf{S} = (\tau \mf{I} + \mathbb{E}_{q(\mf{z})}[\bar{\mf{z}}^\top \bar{\mf{z}}])^{-1}{,\;}
    \mf{m} = \mf{S} \mathbb{E}_{q(\mf{z})}[\bar{\mf{z}}]  \mathbb{E}_{q(\mf{g})}[\mf{g}]
\end{align}

Importantly, the variational mean-field closed-form update for the posterior topic assignment $z_{ij}=k$ depends on both the categorical likelihood of the ICD-9 code and the predictive likelihood of the target disease label: $p(\mf{z} \mid \mf{y},\mf{x}) \propto p(\mf{x} \mid \mf{z})p(\mf{y} \mid \mf{z})$. Leveraging the conditional independence of these two likelihoods, we can calculate the closed-form variational inference update for topic $k$ of ICD-9 code $i$ in patient $j$:
\begin{align}\label{eq:supervised_topic_assignment}
    \gamma_{ijk} & \propto  (\alpha_{k}+E_{q(z^{-(i,j)})} [n_{jk}^{-(i,j)}])
    \nonumber \\
    & \frac{\iota_{b_{ij}} + E_{q(z^{-(i,j)})}[m_{kb_{ij}}^{-(i,j)}]}
    {E_{q(z^{-(i,j)})} [m_{k.}^{-(i,j)}] + \sum_t\iota_t}
    \nonumber \\
    & \frac{\zeta_{kx_{ij}}+E_{q(z^{-(i,j)})} [p_{kb_{ij}x_{ij}}^{-(i,j)}]}{E_{q(z^{-(i,j)})} [p_{kb_{ij}.}^{-(i,j)}]+\sum_w\zeta_{kw}} \nonumber \\
    & \exp \left\{\frac{m_k \mathbb{E}_{q(g_j)}[g_j]}{M_j} -
    \frac{1}{2 M_j^2} 
    \left[2\left(m_k\mf{m}^\intercal\bm{\gamma}_{j/i} + \mf{S}_{k}\bm{\gamma}_{j/i}\right)+m^2_k+S_{kk}\right]\right\}
\end{align}
where $\bm{\gamma}_{j/i} = \sum_{m\neq i}^{M_j} \bm{\gamma}_{ij}$ indicates the sum of all terms except for the $i^{th}$ ICD-9 code, and $\mf{S}_k$ is the $k$th row of the covariance matrix $\mf{S}$. All of the variational expectations have closed-form expression conditioned on the other latent variables:
\begin{align}\label{eq:var_n}
    E_{q(\mf{z}^{-(i,j)})} [n_{jk}^{-(i,j)}] &= \sum_{i^{\prime}\neq i}^{M_{j}}\gamma_{i'jk}
    \notag\\
    E_{q(\mf{z}^{-(i,j)})} [m_{b_{ij}k}^{-(i,j)}] &=
    \sum_{j^{\prime}\neq j}^D\sum_i^{M_{j^{\prime}}}
    [b_{i{j^{\prime}}}=b_{ij}]\gamma_{ij'k}
    \notag\\
    E_{q(\mf{z}^{-(i,j)})} [p_{b_{ij}x_{ij}k}^{-(i,j)}] &=
    \sum_{j^{\prime}\neq j}^D\sum_i^{M_{j^{\prime}}}[
    b_{i{j^{\prime}}}=b_{ij}, x_{i{j^{\prime}}}=x_{ij} ]\gamma_{ij'k}
\end{align}
The above inference technique was originally developed only for LDA and named as collapsed variational Bayesian with zero-order Taylor expansion (CVB0) \cite{teh2007collapsed,asuncion2009smoothing}. Here we extended the inference to more general supervised multi-specialist topic models.

The predictive distribution of the target label $y$ is a Bernoulli distribution when using a Gaussian response since the natural parameter $\mf{w}^\intercal \bar{\mf{z}}$ is identical to the mean parameter:
\begin{align}
    p(\mf{y}_\star \mid \bar{\mf{z}}_\star, \mf{y}, \bar{\mf{z}}) = \text{Bernoulli}\Bigg(\mf{y}_\star \mid \Phi\bigg(\frac{\mf{m}^\intercal \bar{\mf{z}}_\star}
    {(1 + \bar{\mf{z}}_\star^\intercal  \mf S \bar{\mf{z}}_\star)^{\frac{1}{2}}}\bigg)\Bigg)
\end{align}
where $\Phi_j = \Phi(-\lambda_j)$ is the cumulative distribution function (CDF) of the standard normal distribution, and  $\bar{\mf{z}}_\star$ and $\mf{y}_\star$ are the average topic counts and the target disease label, respectively.

The overall inference algorithm is summarized as follows:
\begin{enumerate}
    \item Infer topic assignments $\mf{z}_{ij}$ using Eq. \eqref{eq:supervised_topic_assignment}
    \item Update sufficient statistics for the topic inference by Eq. \eqref{eq:var_n}
    \item Update target prediction parameters by Eq~\eqref{eq:trunc_norm} and Eq~\eqref{eq:para_update_w}
    \item Calculate the expectation of the target $y$ (\textbf{Appendix}~\ref{appendix:expectation_calculation})
    \item Update the Dirichlet hyperparameters (\textbf{Appendix}~\ref{sec:appendinx_hyperparameters})
    \item Repeat step 1-5 until little change in ELBO (default: 1e-6).
\end{enumerate}
For efficient inference over large-scale EHR data, we employed a stochastic collapsed variational inference (SCVB0) \cite{hoffman13a,Li:2020exa}. The full details are described in \textbf{Appendix} (Section \ref{sec:appendix:derivation}).

\section{Data}\label{sec:data}

\subsection{Quebec CHD Database (154,775 patients)}\label{sec:chd_data}
In Quebec (Canada) where universal health care is provided, every resident is assigned a unique Medicare number and all health services rendered are systematically recorded until death. In this study, three EHR databases were merged by the unique patient medicare numbers: (1) the physician’s services and claims database from l983 to 2010; (2) the hospital discharge summary database from l987 to 2010; (3) the vital status database from 1983 to 2010. As a result, we have collected the EHR data for 154,775 patients. Demographic information including age and sex were also included in the databases and no specific age or sex biases were observed. The study population included patients with at least one CHD-related ICD-9 diagnosis between 1983 and 2010, whose ICD diagnoses were made by one of the 48 CV or non-CV specialists (e.g., cardiologist, thoracic surgeon, cardiac surgeon, cardio-vascular and thoracic surgeon, etc). In total, there are 9373 unique ICD-9 codes. The gold-standard labels are the binary label of CHD diagnosis made by clinicians after careful manual auditing the patient records using a clinician-developed rule-based algorithm. Among the 154,775 patients, 84,498 patients were diagnosed as true CHD patients ($y_j=1$) and the rest are non-CHD patients ($y_j=0$). Therefore, if we were to only use the CHD-related ICD-9 code to predict CHD labels, we could have only achieved an accuracy of 54.5\%.

\subsection{PopHR database} \label{sec:PopHR_data}
The Population Health Record (PopHR) is a semantic web application for measuring and monitoring population
health and health system performance \cite{Powell:2017bd}. 
The public health insurance provider in Quebec, Canada (Régie de l’assurance maladie du Québec, RAMQ) provided the data on health service use. PopHR’s current version uses an open, dynamic cohort of approximately 1.4 million people, created by capturing a 25\% random sample in the census metropolitan area of Montreal between 1998 and 2014. Follow-up ended when people died or moved out of the region of Montreal. The administrative database includes outpatient diagnoses and procedures submitted through billing claims to  RAMQ, and procedures and diagnoses from hospital records. Drug dispensation data are available for people who have drug insurance through RAMQ (which includes approximately half the population and all those over 65 years of age). All data are linked through an anonymized version of the RAMQ identification number. As a proof-of-concept, we focused on a subset of the patient cohort with two distinct target diseases as described below.

\paragraph{Chronic Obstructive Pulmonary Disease (COPD)}
Our goal in this application is to infer specialist-dependent COPD topics and predict the true COPD patients. To this end, we retrieved 73,791 patients with at least one COPD related ICD-9 code (e.g.  491x, 492x, 496x) from the physician billings table of the PopHR database. The gold-standard labels were assigned to patients as of their incident event (COPD diagnostic code for hospitalization or medical billing) occurring after a minimum of two years of time at risk \cite{verma2018modeling}. Among the 73,791 patients identified through ICD-9 codes, 67,380 patients were confirmed to be the true COPD patients and the rest are non-COPD patients. In total, there are 6,625 unique ICD-9 codes and 43 unique specialists.

\paragraph{Insulin usage among diabetes patients}
In this application, we aim to infer latent topics that are predictive of whether each diabetes patient started the insulin medication 6 months after their initial diagnosis. This is useful in forecasting the disease exacerbation. We extracted 78,712 diabetes patients with the ICD-9 codes 250x and continuous public drug insurance for hospitalization or medical billing. We set the first diabetes diagnosis of a patient as the start of insurance coverage. We selected patients with a continuous public drug insurance after the first diagnosis of diabetes-related code, with continuous insurance defined as: (1) patients with at least 6 months of uninterrupted insurance, or; (2) patients with interrupted insurance records for which interruptions are less than 2 months. To avoid misclassifying the first dispensation in patients with an unrecorded history of insulin, we removed patients who used insulin within 6 months after the first diagnosis of a diabetes-related code. 

For the remaining patients, we only used their ICD codes observed accumulatively up to the first diagnosis to predict their future drug usage. We treated a patient as positive if he or she started to use insulin 6 months after being diagnosed with diabetes. Among the 78,712 diabetes patients, 11,433 patients were labeled as insulin users. In order to evaluate our model on a balanced dataset, we randomly sampled 11,433 negative patients to obtain a perfectly balanced dataset (50\% positive patients, 50\% negative patients). There are 5,477 unique ICD-9 codes and 43 specialists in the resulting balanced diabetes dataset.

\section{Experiments}
We sought to evaluate both model interpretability and prediction accuracy of our \mxr~model. Because the true topics of real dataset are not known, we first used simulation to assess whether \mxr~can recapitulate the ground-truth topics. We used the generative process of the proposed model to obtain sample data. We simulated 2,500 patients, 750 ICD diagnosis codes and 48 specialists for the evaluation. We considered 25 topics to model toy data with a 80/20 train-test split. For the purpose of comparison, we used LDA as the baseline model \cite{blei2003latent}. We ran LDA implemented in Python package scikit-learn with default variational inference algorithm. We evaluated disease target prediction by the receiver operating characteristic (ROC) curve and precision-recall curve (PRC). We directly predicted the disease target labels using \mxr. For the unsupervised model LDA, we trained a separate Bayesian logistic regression to predict labels given the LDA-inferred patient topic mixtures. Overall, \mxr~achieved an excellent topic recovery and outperformed LDA (Fig. \ref{fig:quantitative_result_simulation}a). Details were described in \textbf{Appendix}~\ref{sec:simulation_results}. 

We then evaluated our \mxr~model based on the real EHR data from Quebec CHD, PopHR-COPD, and PopHR-diabetes datasets (Section \ref{sec:data}). As a baseline method, we evaluated MixEHR \cite{Li:2020exa}, which is an unsupervised multi-modal topic model. We also  evaluated LDA and the supervised LDA (sLDA), both of which operate on flatten ICD-9 codes as single category \cite{blei2003latent,mcauliffe2008supervised}. We applied the Bayesian Logistic Regression classifier that uses the topic mixtures inferred by MixEHR or LDA to predict labels. We also compared our approach with supervised models including Least Absolute Shrinkage and Selection Operator (LASSO) \cite{Tibshirani1996RegressionSA}, Random Forest (RF), Gradient Boosting (GB), and two-layer feedforward neural network (NN), which directly use ICD-9 codes as raw features to predict CHD labels. LDA, LASSO, RF, and GB were implemented by the Python package scikit-learn with the default optimization algorithms. NN with two fully connected layers and 100 hidden units per layer was implemented in PyTorch. 

Additionally, we evaluated several state-of-the-art EHR-focused models including MixEHR \cite{Li:2020exa}, Deep Patient (DP)  \cite{Miotto2016-hn} (\url{https://github.com/natoromano/deep-patient})  , Graph-based Attention Model (GRAM) \cite{Choi:2017fm} (\url{https://github.com/mp2893/gram}), and Sparsity-inducing Collected Non-negative Matrix Factorization (SiCNMF) \cite{Gunasekar2016-gr} (\url{https://github.com/sgunasekar/SiCNMF}) using available published codes from the corresponding GitHub repositories.

For all three applications, we split each dataset into 70\% training, 10\% validation, and 20\% testing sets. For the topic models (i.e., MixEHR, LDA, sLDA, and our \mxr), we used the validation set to choose the best number of topics based on the unsupervised perplexity (i.e., the negative held-out log-likelihood) on the the 10\% validation patients. We evaluated on the 20\% test patients in predicting the target disease based on the area under the ROC curves (AUROC) and precision-recall curves (AUPRC). To obtain robust prediction estimates, we conducted 10 repeated runs with random 70\%-train/10\%-validation/20\%-test splits and recorded the mean values and standard deviations of AUROC and AUPRC for each model on each test set in Table.\ref{table: prediction_table}. The detailed model specifications and hyperparameters were described in Appendix \ref{appendix:model_specifications}.

\begin{figure*}[!t]
\centering\includegraphics[width=0.8\textwidth]
{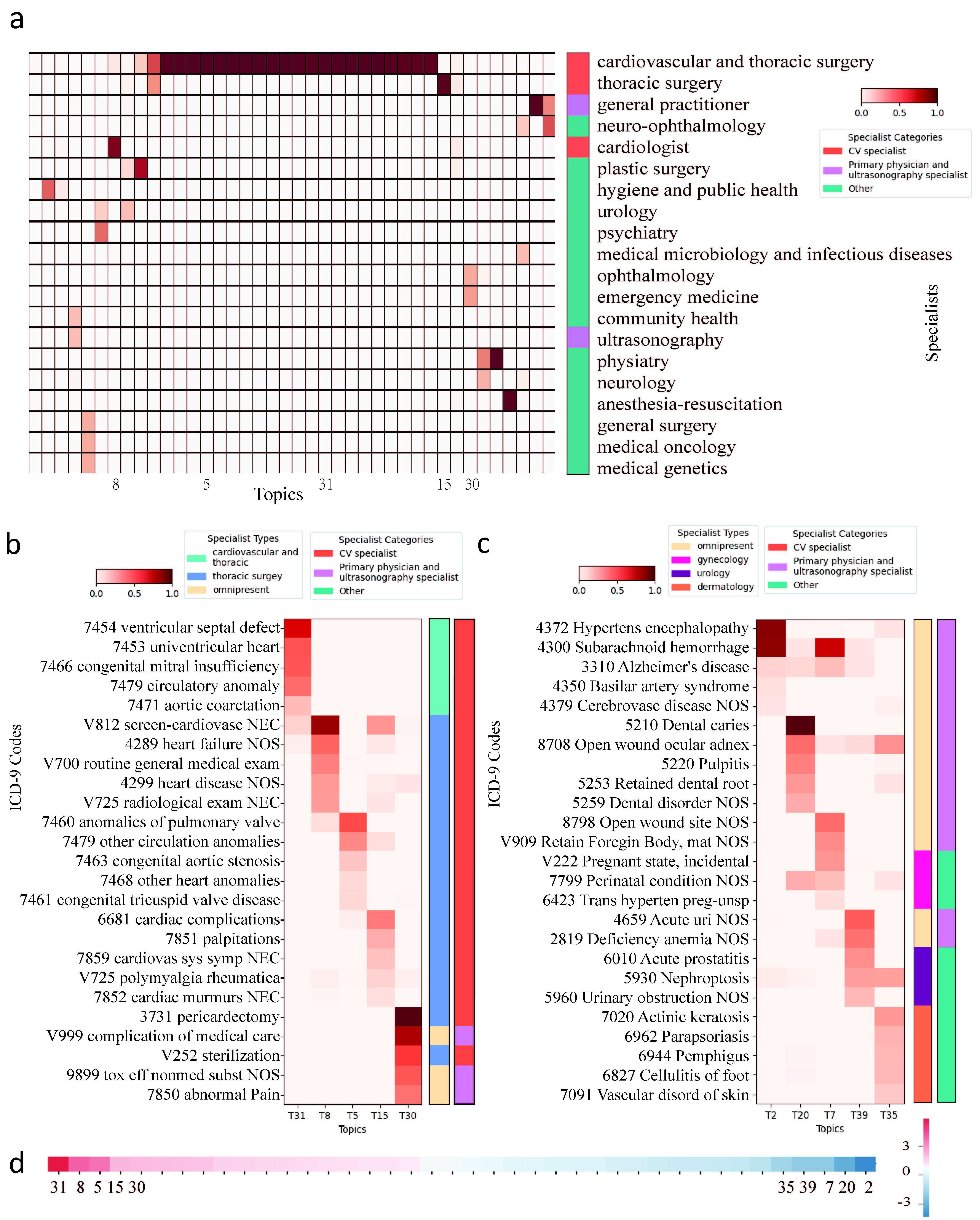}
\caption{Disease topics inferred by \mxr~on the CHD dataset. (a) The inferred specialist topics. The color intensities are proportional to the inferred probabilities of specialists under each topic and the side bar indicates the specialist categories. The topics that are numbered are the most predictive topics of CHD. The full specialists are illustrated in Fig. \ref{fig:specialist_topic_CHD}. (b) and (c) Top ICD-9 codes from the five most positively predictive topics and the five most negatively predictive topics. The side bars indicate the specialist types and categories. (d). Linear coefficients for the 40 topics in decreasing order.}
\label{fig:topic_modelling_CHD}
\end{figure*}

\begin{figure*}[t]
\centering\includegraphics[width=0.9\textwidth]{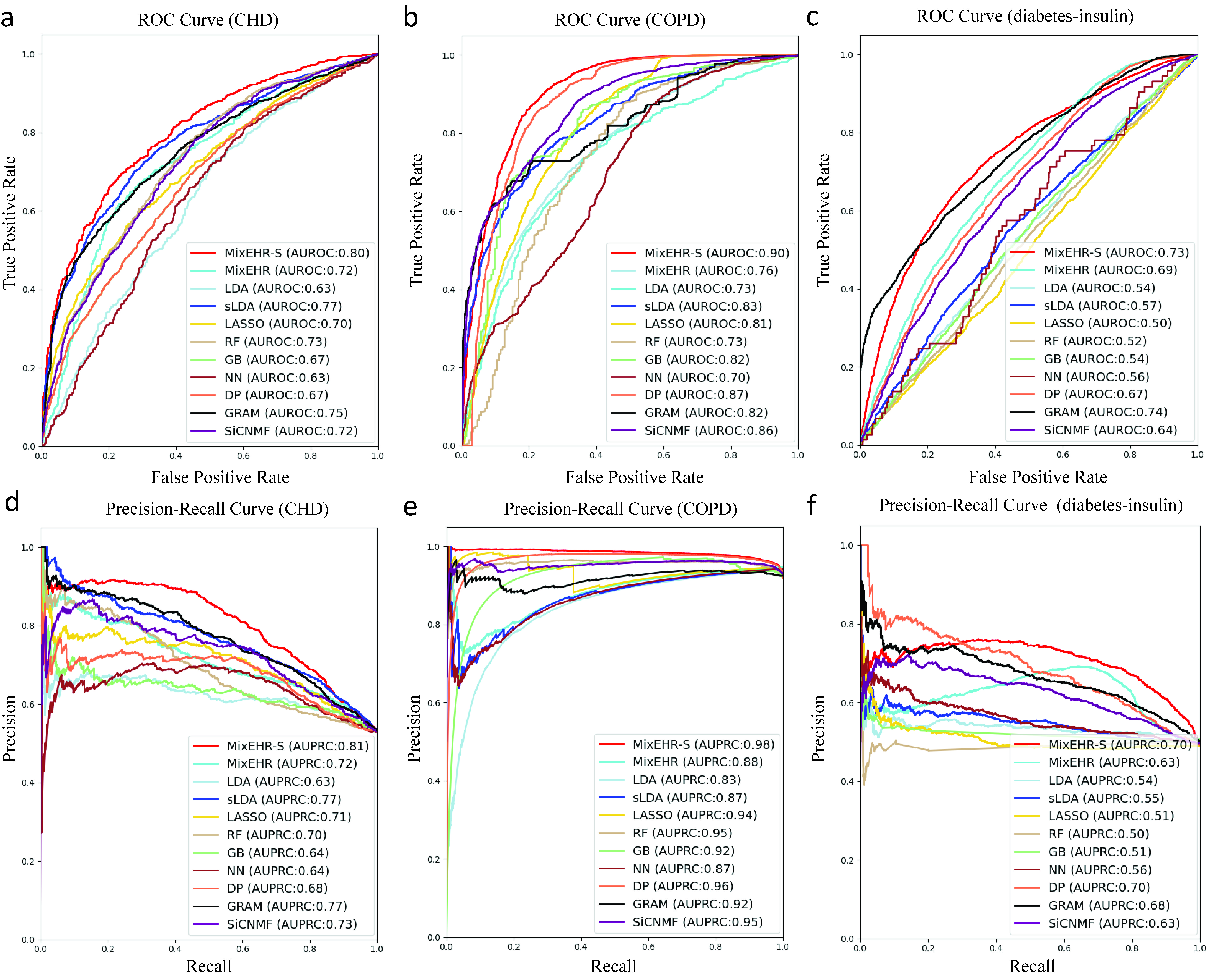}
\caption{Prediction accuracy of responses comparing \mxr~model, the topic models (MixEHR, LDA, and sLDA), the models which directly apply on raw ICD-9 codes (LASSO, RF, GB, and NN), and the EHR-focused models (DP, GRAM, and SiCNMF). \textbf{a-c}  (\textbf{d-f}) ROC and PR curves of all methods on CHD, COPD, and diabetes-insulin predictions with AUROC (AUPRC) of each method indicated as inset in each panel.
}
\label{fig:prediction}
\end{figure*}

\begin{figure*}[!t]
\centering\includegraphics[width=\textwidth]
{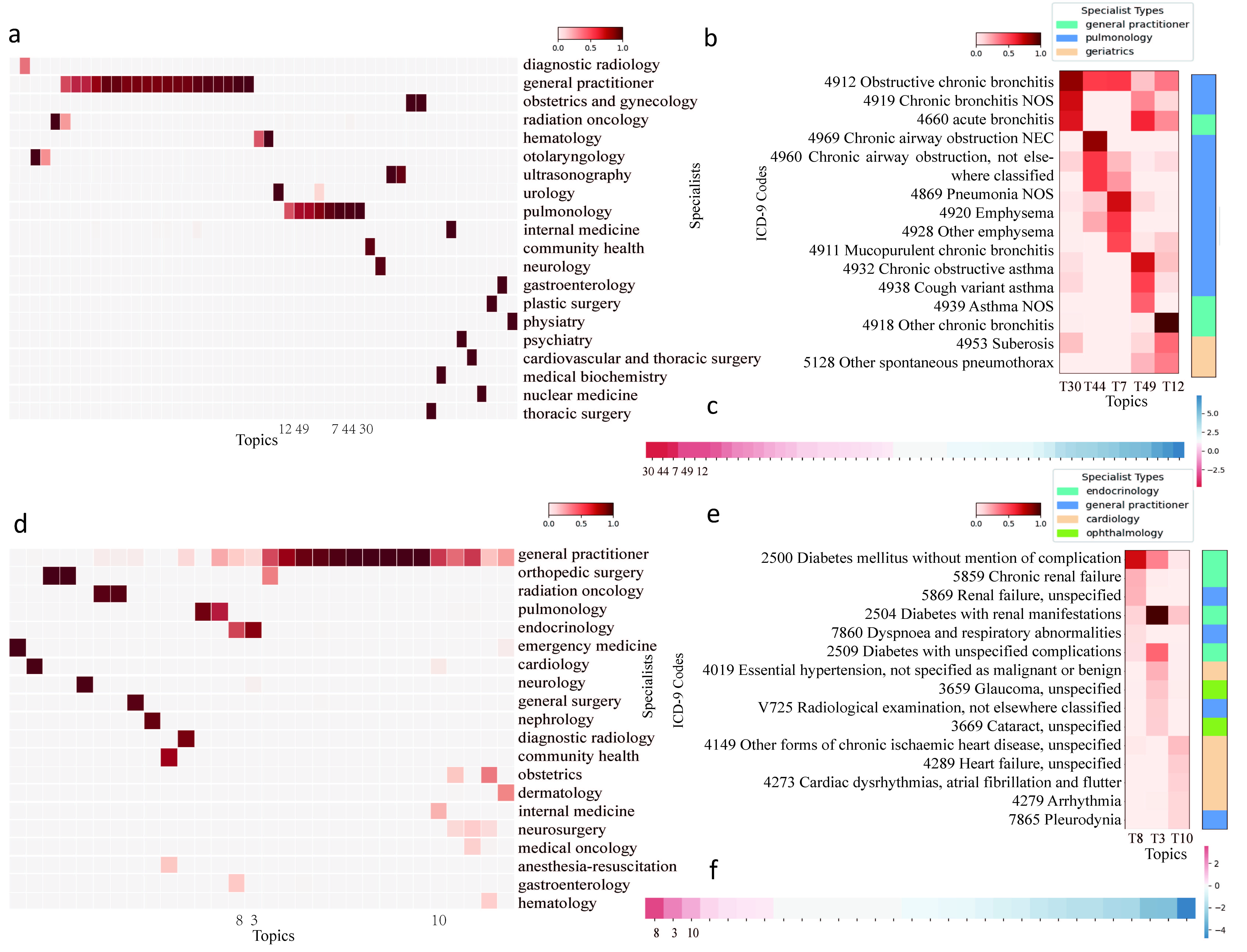}
\caption{Disease topics inferred from the COPD and diabetes patients' EHR data extracted from the PopHR database. 
\textbf{a}. Specialist topics. Same as in Figure \ref{fig:topic_modelling_CHD}, the color intensities are proportional to the probabilities of specialists under each topic and the side bar indicates the specialist categories. 
\textbf{b}. Top three ICD-9 codes from the five most positively predictive topics. The side bars indicate the specialist types. \textbf{c}. Linear coefficients for the 50 topics in decreasing order. \textbf{d-f}. specialist topics, top ICD-9 codes per top topic, and linear coefficients for the 6-month insulin-usage prediction among the diabetic patients.
}
\label{fig:topic_modelling_COPD_diabetes}
\end{figure*}

\section{Results}
\subsection{Inferring specialist-dependent topics from Quebec CHD Dataset}
We investigated the inferred 40-topic mixtures of 48 specialists that give the lowest perplexity on the validation set (Fig.~\ref{fig:topic_modelling_CHD}a) (i.e., $\bm{\beta}$). As expected, we observe high probabilities for CV specialists, such as cardiovascular and thoracic surgery. Based on the learned topic coefficient $\mf{w}$, topics T31 and T8 are the two most predictive topics for CHD and are strongly associated with cardiovascular and thoracic surgery and cardiologist, respectively. For each topic, we chose the top five ICD-9 codes to reveal its meaning. We found a strong connection between the inferred CHD-positive topics and the disease pathology. In particular, topics T31, T8, and T5 contain many CHD-related diagnosis codes such as ventricular septal defect (7454), univentricular heart (7453), and anomalies of pulmonary valve (7460), acute conditions such as heart failure (4289) under T5, and procedure code such as cardiac examination under T5. Moreover, all of the top ICD diagnosis codes under topic T31 and T5 come from the CV specialists (i.e. cardiovascular and thoracic surgery or thoracic surgeon). Also, the top ICD-9 codes under topic T15 such as cardiac complications and cardiovascular symptoms also mostly make clinical sense. Topic T30, which is weakly associated with heart related surgery, gives less interpretable results since several of its top codes are diagnosed by the non-CV specialists. 

Interestingly, the selected negatively predictive topics are also clinically coherent although all of them are not related to CHD as expected. For instance, we observed a clear enrichment for cerebrovascular diseases in T2. Topic T20 represents dental diseases. All of the top ICD-9 codes under the negatively predictive topics come from non-CV practitioners. Therefore, our inferred topics could be used as a clinically relevant departure point to uncover more specific therapeutic information.

We then examined the disease topic mixture memberships along the patients dimension. We selected 5 patients with
the highest proportions for the top 5 most predictive topics (Fig.~\ref{fig:patient_topic_plot}b), who we considered as high-risk CHD patients. The majority of the top patients are indeed true CHD patients. In contrast, most patients under the top 5 negatively predictive topics do not have CHD labels (Fig.~\ref{fig:patient_topic_plot}c). We then examined the 5 highest-scoring ICD-9 diagnosis codes among these patients under the five most positively predictive topics and the five most negatively predictive topics. As expected, the top patients in the positively predictive topics were mostly diagnosed with the CHD-related ICD-9 codes. In particular, most of the patients under topic T31 have highly CHD-specific ICD-9 codes 7454, 7453 and 7471, corresponding to the diagnoses of ventricular septal defect, univentricular heart, and aortic coarctation, respectively. Some of the top patients were not confirmed with CHD and may be deemed as the high-potential CHD patients. Some of the top 50 patients under topic T31 have neither the CHD label nor the top 5 CHD-related ICD codes. Nonetheless, these patients possess the other top CHD-related ICD-9 codes under the topic T31 (Fig.~\ref{fig:patient_topic_plot}d).  

\begin{table}[!h] 
\fontsize{7.5}{8}\selectfont 
\caption{Prediction performance of \mxr~and other methods on CHD, COPD, and diabete-insulin predictions. The mean values and standard deviations (in the brackets) of AUROC and AUPRC for each model  each test set are computed on 10 randomly training and testing splits. For each application, the highest average AUROC and AUPRC among all methods are in bold.} 
\label{table: prediction_table}
\begin{center}
\begin{tabular}{| l | c | c | c | c | c | c |}
\hline
\multirow{2}{*}{Method}  
& \multicolumn{2}{l|}{CHD}  
& \multicolumn{2}{l|}{COPD}  
& \multicolumn{2}{l|}{diabetes-insulin}  
\\
\cline{2-7}
 & AUROC & AUPRC & AUROC & AUPRC & AUROC & AUPRC \\
\hline 
\textbf{MixEHR-S} & \textbf{0.8007} & \textbf{0.8122} & \textbf{0.9036} & \textbf{0.9774} & \textbf{0.7341} & \textbf{0.7006} \\
 & (0.0263) & (0.0228) & (0.0290) & (0.0122) & (0.0469) & (0.0376) \\
\hline 
\textbf{MixEHR} & 0.7214 & 0.7167 & 0.7621 & 0.8804 &  0.6892 & 0.6346 \\ 
 & (0.0384) & (0.0462) & (0.0429) & (0.0607) & (0.0463) & (0.0471) \\
\hline 
\textbf{LDA} & 0.6292 & 0.6311 & 0.7285 & 0.8318 & 0.5418 & 0.5385\\ 
 & (0.0327) & (0.0341) & (0.0362) & (0.0686) & (0.0210) & (0.0188) \\
\hline 
\textbf{sLDA} & 0.7684 & 0.7708 & 0.8317 & 0.8665 & 0.5622 & 0.5509 \\
 & (0.0258) & (0.0272) & (0.0362) & (0.0471) & (0.0124) & (0.0146) \\
\hline 
\textbf{LASSO} & 0.6967 & 0.7130 & 0.8125 & 0.9382 & 0.5017 & 0.5083 \\ 
 & (0.0381) & (0.0392) & (0.0431) & (0.0271) & (0.0109) & (0.0086) \\
\hline 
\textbf{RF} & 0.7291 & 0.7012 & 0.7312 & 0.9453 & 0.5238 & 0.5044 \\
 & (0.0131) & (0.0155) & (0.0181) & (0.0093) & (0.0113) & (0.0081) \\
\hline 
\textbf{GB} & 0.6703 & 0.6447 & 0.8239 & 0.9164 & 0.5398 & 0.5124 \\ 
 & (0.0107) & (0.0142) & (0.0154) & (0.0142) & (0.0220) & (0.0093) \\
\hline 
\textbf{NN} & 0.6334 & 0.6409 & 0.6976 & 0.8708 & 0.5588 & 0.5620 \\ 
 & (0.0439) & (0.0504) & (0.0360) & (0.0557) & (0.0591) & (0.0475) \\
\hline 
\textbf{DP} & 0.6675 & 0.6825 & 0.8655 & 0.9634 & 0.6701 & \textbf{0.7028} \\ 
 & (0.0305) & (0.0322) & (0.0206) & (0.0125) & (0.0256) & (0.0296) \\
\hline 
\textbf{GRAM} & 0.7493 & 0.7687 & 0.8245 & 0.9239 & \textbf{0.7387} & 0.6815 \\ 
 & (0.0346) & (0.0221) & (0.0168) & (0.0102) & (0.0438) & (0.0399) \\
\hline 
\textbf{SiCNMF} & 0.7241 & 0.7313 & 0.8591 & 0.9546 & 0.6432 & 0.6283 \\
 & (0.0436) & (0.0387) & (0.0260) & (0.0168) & (0.0423) & (0.0368) \\
\toprule
\end{tabular}
\end{center}
\end{table}

\subsection{CHD target label prediction}
Quantitatively, \mxr~conferred the highest AUROC (80.07\%) and AUPRC (81.22\%) among all methods (Fig. \ref{fig:prediction}a and d, Table.\ref{table: prediction_table}). The LDA model that ignore distinct specialists performed a lot worse while the supervised sLDA model was a slightly better. \mxr~also outperformed the unsupervised variant MixEHR, which achieved 72.14\% AUROC and 71.67\% AUPRC. Additionally, the baseline discriminative models LASSO, RF, GB, and NN, which directly use the raw features, also did not predict CHD outcomes as accurately as our \mxr. This is due to their inability to model the distribution of the EHR input data. As a result, they are more sensitive to the sparsity and noise intrinsic to the EHR data. 

Compared to the above baseline models, the existing EHR-focused models demonstrated generally better prediction performance. DP conferred relatively low prediction accuracy possibly due to its need to operate on extensively prepossessed and filtered EHR codes. GRAM utilized the ICD-related knowledge graph to embed ICD code, which helped achieving more accurate prediction (74.93\% AUROC and 76.87\% AUPRC). Together, we attribute the highest performance of our \mxr~to its simultaneous inference of the specialist-specific topics and the predictive topic coefficients. 

\subsection{Modeling PopHR-COPD data}

We then examined the 50-topic mixtures over the 43 specialists that we inferred from the PopHR-COPD data (Section \ref{sec:PopHR_data}). As in the CHD topic analysis above, we focused on the 5 most predictive topics of COPD by examining the specialist distribution  (Fig.~\ref{fig:topic_modelling_COPD_diabetes}a, c). Indeed, we observe that the top 5 topics (T30, T44, T7, T49, T12) are strongly associated with pulmonology specialist, which is closely related to the COPD diagnosis. Consistent to the trend in the CHD analysis, the disease relevance increases as the topic's predictive coefficient increases (Fig.~\ref{fig:topic_modelling_COPD_diabetes}c; \textbf{Appendix}~\ref{sec:complte_figures}). In particular, T30 is the most predictive one among all 50 topics, which exhibit the highest probability for pulmonology specialist. Topics T7 and T12 exhibit relatively weaker associations with pulmonology specialist and therefore weaker effect sizes for COPD.


The top 3 ICD-9 codes under the most predictive topic T30 are all closely related to COPD: Obstructive chronic bronchitis(4912), Chronic bronchitis NOS (4919), and acute bronchitis (4660) (Fig.~\ref{fig:topic_modelling_COPD_diabetes}b). The top second and top third topics T44 and T7 also contain multiple COPD-related codes, which represent Chronic airway obstruction and emphysema, respectively. Overall, 10 out of the 15 top ICD-9 codes are from the pulmonology specialist type and the rest of the codes are from general practitioner and geriatrics. This makes sense since pulmonology specialist is more likely to make diagnosis for these COPD-related ICD codes. Interestingly, although topics T49 and T12 are less predictive of COPD, they are strongly associated with asthma and lung malfunction, which often manifest similar symptoms as COPD.

We then evaluated the prediction accuracy (Fig.~\ref{fig:prediction}b and e, Table.\ref{table: prediction_table}). \mxr~achieves the highest AUROC of 90.36\% among all methods. The other top performing models namely DP, SiCNMF, and GRAM conferred AUROC 86.55\%, 85.91\%, and 82.45\%, respectively. LDA alone did not work well with only 72.85\% AUROC, possibly due to the its inability to capture the predictive topics of COPD and its negligence of the multi-specialist ICD-9 distributions.

Compared with LDA, the unsupervised MixEHR model \cite{Li:2020exa} that learns specialist specific topics achieved higher AUROC of 76.21\%. However, it still fell a lot short on the prediction task compared to \mxr. We observed much higher accuracy by the supervised LDA (sLDA) with 83.17\% AUROC. Therefore, adding the supervised component into the model can improve predictive performance. 

The discriminative models namely LASSO, RF, GB, NN achieved a diverse performances ranging from the lowest 69.76\% (LASSO) to 82.39\% (GB). LASSO failed to work on the raw ICD-9 codes with only 73.12\% AUROC, implying that it is inadequate to place only sparse constraints on the otherwise high-dimensional, highly sparse, and highly noisy EHR code features. Most methods achieve high AUPRC (Fig. \ref{fig:prediction}e). This is partially due to the highly unbalanced nature of dataset (approximately 92\% of observations are from positive class).

\subsection{Predicting insulin usage from PopHR-Diabetes patients}

Finally, we applied \mxr~to predict insulin usage among the over 78000 confirmed diabetes patients (Section \ref{sec:PopHR_data}). As the analysis above, we first qualitatively examined most predictive topics of insulin usage (Fig. \ref{fig:topic_modelling_COPD_diabetes}d,e). Here we focused on only the top 3 topics because of the rapid decrease of the predictive coefficients following them (Fig. \ref{fig:topic_modelling_COPD_diabetes}f). The most predictive topics T8 and T3 are asscoiated with the endocrinology and cardiology specialist types, which have strong association with diabetes diagnosis, whereas the top third topic T10 is connected with internal medicine and gastroenterology (Fig. \ref{fig:topic_modelling_COPD_diabetes}d; \textbf{Appendix}~\ref{sec:complte_figures}).

The top ICD-9 codes under topics T8 and T3 are also highly enriched for three diabetes codes, namely 2500 diabetes mellitus without mention of complication, 2504 diabetes with renal manifestations, and 2509 diabetes with unspecified complications. We observed that the two renal failure ICDs under T8 with modest probability and interpreted them as renal complications of diabetes. Interestingly, topic T10 contains multiple ICD codes on cardiovascular conditions (ischemic heart disease, heart failure), which are diagnosed by cardiology specialists. Therefore, it is likely that topic T10 characterizes the cardiovascular complications of diabetes.

We then predicted the insulin usage outcome 6 months after the first diagnosis of diabetes (Fig.~\ref{fig:prediction}c and f, Table.\ref{table: prediction_table}). \mxr~achieved the highest AUROC (73.41\%) and AUPRC (70.06\%) or largely on-par with GRAM and DP, respectively. Other EHR-based models fell short. The discriminative models LASSO, RF, GB, and NN performed quite poorly with AUROC and AUPRC below 60\%. Among the topic models, MixEHR outperforms LDA and sLDA, confirming the benefits of learning the distribution of specialist-specific topics. Together, with \mxr, we demonstrate the advantages of heterogeneous and and task-dependent topic modelling of the EHR data.


\section{Discussion}
In this study, we presented \mxr~as an extension of our MixEHR \cite{Li:2020exa}. Compared to the existing methods, we explicitly model the specialist assignments and ICD-coded diagnoses as latent topics. \mxr~can simultaneously infer the distribution of an arbitrary number of patient-specialist assignments and predict a binary disease label based on the learned disease topics. Compared to the simpler topic models, the relatively higher model complexity of our \mxr~does not actually incur higher computational costs especially when modeling large-scale EHR data with up to 150,000 patients. This is attributable to our closed-form mean-field variational expectation-maximization updates and our collapsed stochastic variational inference algorithm. We demonstrated \mxr~through a comprehensive set of experiments on both simulated data and two large-scale EHR databases with three targeted real-world applications. Consistent throughout our experiments, \mxr~not only infers meaningful specialist-dependent topics but also makes accurate prediction of disease target labels.

Besides the three applications, \mxr~can generalize to other EHR data with non-randomly distributed heterogeneous data types. For example, patients with brain disorder are more likely to have electroencephalography data, and patients with pulmonary problems will more likely to have chest radiograph. In other words, the specific data types observed among patients depend on the patient disease types.

In our future work, we will extend \mxr~for multi-class prediction model. This will allow us to model more heterogeneous patient cohort with  related target diseases. EHR data are longitudinal. Our current \mxr~requires aggregating patient diagnoses over time to have a single collapsed data point to represent each patient. As one of our ongoing projects, we are exploring dynamic topic model \cite{Lafferty:2015wj} to properly account for time-dependent EHR data.

\section*{Funding}
YL is supported by Microsoft Research. AM is supported by Canadian Institute of Health Research (CIHR) Foundation Grant.

\bibliographystyle{plain}
\bibliography{sample-authordraft}

\onecolumn
\appendix

\setcounter{table}{0}
\renewcommand{\thetable}{S\arabic{table}}%
\setcounter{figure}{0}
\renewcommand{\thefigure}{S\arabic{figure}}
\setcounter{page}{1}%

\section{Supplementary Information}

\subsection{Description of variables of Graphical Model}

\begin{table}[!htbp] 
\fontsize{11}{9}\selectfont 
\caption{Description of Variables of \mxr} 
\label{table: vardescription}
\begin{center}
\begin{tabular}{| l | l |}
\hline \hline
\textbf{\normalsize Variable}  & \textbf{\normalsize Definition}  \\
 \hline 
\textbf{$y_j$} & Response connected to patient $j$ \\ \hline 
\textbf{$x_{ij}$}  & ICD-9 code $i$ of patient $j$ \\ \hline  
\textbf{$b_{ij}$} & Specialist type for ICD-9 code $i$ of patient $j$ \\ \hline 
\textbf{$z_{ij}$} & Topic assignment for ICD code-9 $i$ of patient $j$ \\ \hline 
\textbf{$\theta_{j}$}& Topic mixture of patient $j$ \\ \hline 
\textbf{$\beta_{k}$} & Specialist mixture given topic $k$ \\ \hline 
\textbf{$\eta_{kt}$} & ICD-9 code mixture given topic $k$ and specialist $t$ \\ \hline 

\textbf{$\alpha$} & Dirichlet hyperparamter \\ \hline 
\textbf{$\iota$} & Dirichlet hyperparamter \\ \hline 
\textbf{$\zeta_k$} & Dirichlet hyperparamter \\ \hline

\textbf{$g_j$} &  Latent disease liability of patient $j$ \\ \hline 
\textbf{$w$} &  Linear coefficients \\ \hline
\textbf{$\tau$} & Precision variable of Gaussian distribution for regression coefficient $w$ \\ \hline 

\toprule
\end{tabular}
\end{center}
\end{table}

\subsection{Simulation Results}\label{sec:simulation_results}

\begin{figure}
\centering\includegraphics[width=\linewidth]{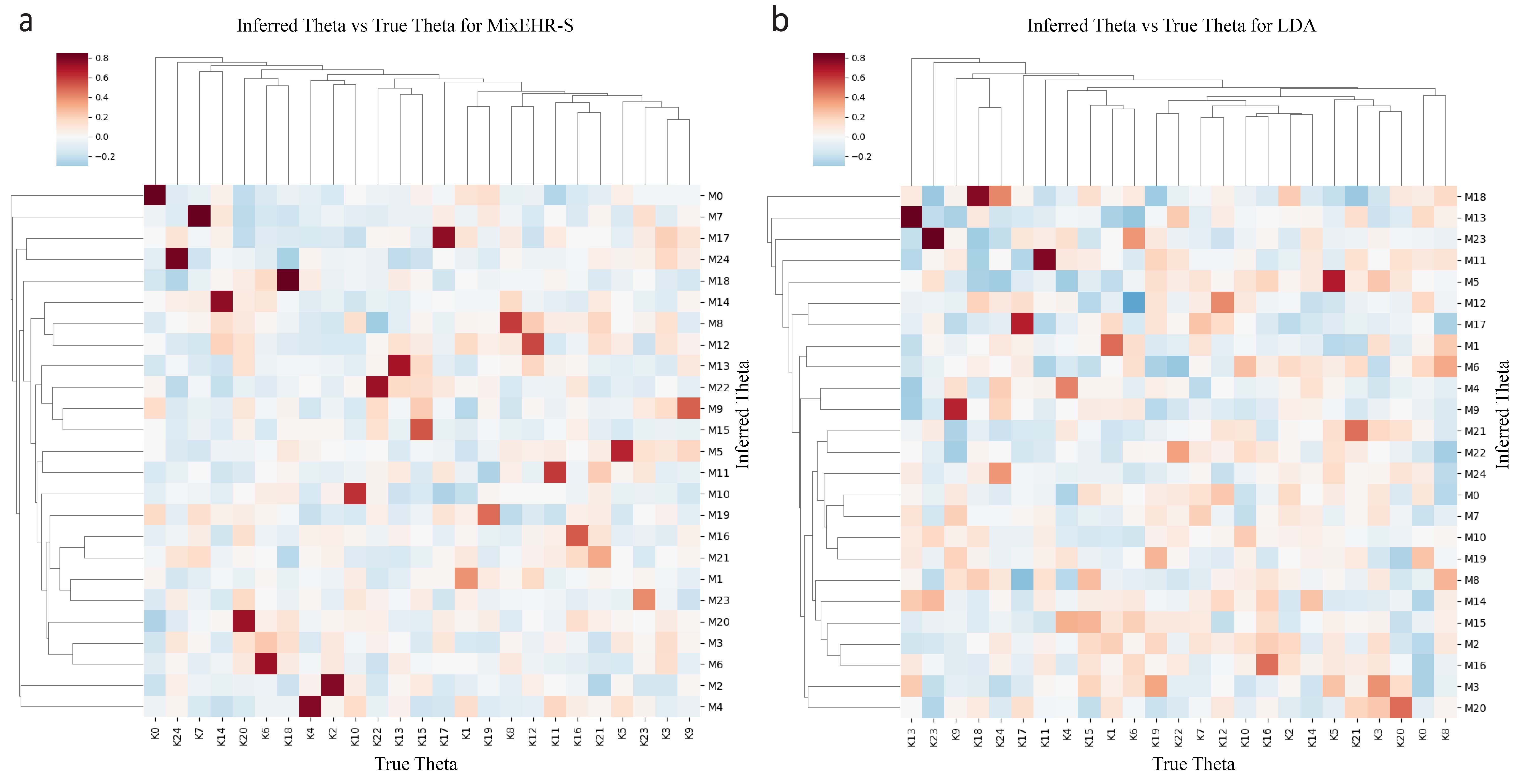}
\caption{Topic recovery comparison from simulation dataset. To assess whether our model can recapitulate the groundtruth topics, we correlate every inferred topic with every groundtruth topic. In particular, we correlate the $D \times K$ inferred patient topic mixture $\hat{\bm{\theta}}$ for $D$ patients and $K$ topics with the groundtruth $D \times K$ patient topic mixture matrix. The resulting Pearson correlation is therefore a $K \times K$ symmetric matrix. \textbf{a} Topic correlation using the inferred topic mixture by \mxr~model. \textbf{b} Topic correlation using the inferred topic mixture by LDA.}
\label{fig:topic_recovery_simulation}
\end{figure}

\begin{figure}
\centering\includegraphics[width=\linewidth]
{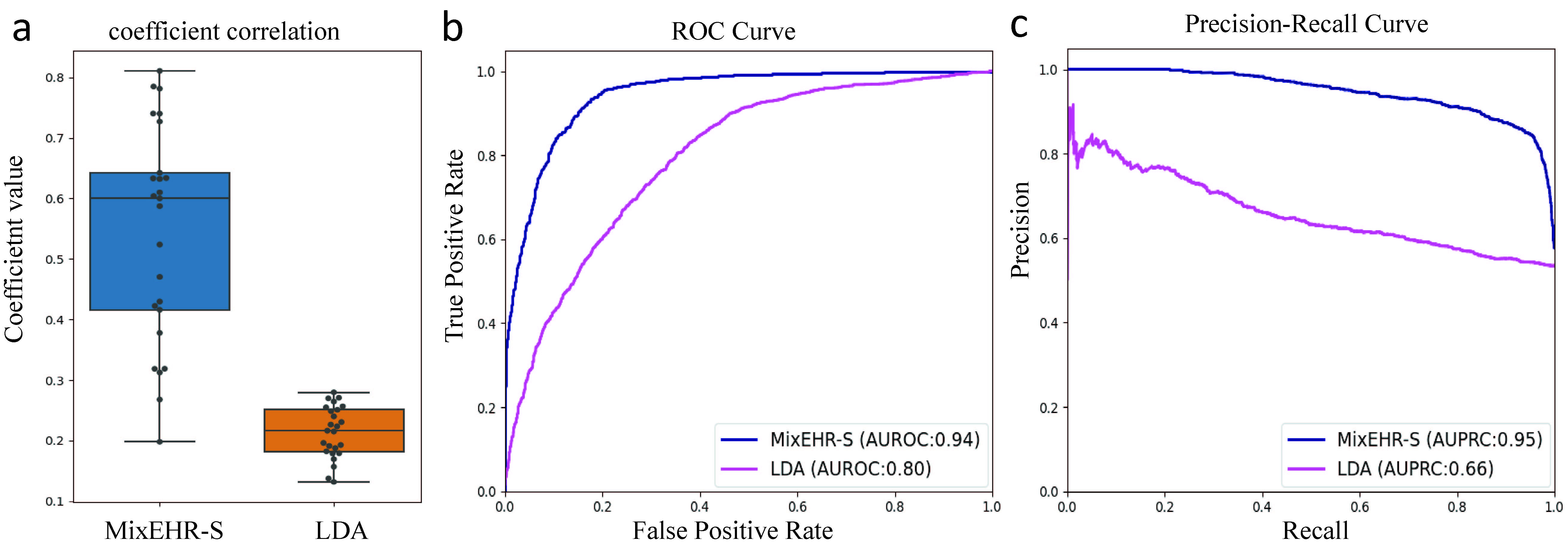}
\caption{Evaluation of simulation data. \textbf{a} Correlation between true topics and inferred topics by \mxr~and LDA. Response prediction evaluated by \textbf{b} ROC and \textbf{c} PRC.}
\label{fig:quantitative_result_simulation}
\end{figure}

For topic modeling, we quantitatively assessed the correlation between the inferred topics and the ground truth topics (Fig.\ref{fig:topic_recovery_simulation}). We computed the Pearson correlation between the inferred patient topic mixtures and the patient topic mixture matrix. \mxr~achieved an excellent topic recovery and outperformed LDA (Fig. \ref{fig:quantitative_result_simulation}a). The improvement of \mxr~over LDA is attributable to explicitly modeling specialist topic distributions. In particular, LDA does not infer the specialist-patient assignments and multi-specialist topic distribution. For the target disease prediction, \mxr~achieved 93.54\% AUROC and 95.26\% AUPRC whereas LDA obtained only 80.17\% and 66.41\% AUROC and AUPRC, respectively (Fig.~\ref{fig:quantitative_result_simulation}b,c). The results therefore support the benefits of jointly modeling of multi-specialist topics and predicting labels compared to the LDA + Logistic Regression pipeline approach.

\subsection{Baseline methods}
\label{appendix:model_specifications}

We used the validation sets of the 3 datasets to tune the number of latent topics for topic-based models in terms of perplexity (with the chosen topic numbers in the brackets): MixEHR ($K$=40, 45, and 25 for CHD, COPD, diabetes-insulin, respectively);  LDA ($K$=30, 35, and 20, respectively); sLDA ($K$=25, 35, and 20, respectively). 

We also implemented machine learning algorithms that learn raw EHR data by using Python packages scikit-learn and PyTorch. For LASSO, we obtained best penalty parameter for the L1 term using grid search in scikit-learn package (\url{https://scikit-learn.org/stable/modules/generated/sklearn.model_selection.GridSearchCV.html}). We set the number of decision trees and max depth as 300 and 3 for RF classifier. Besides, we chose the number of boosting stages and learning rate as 300 and 0.1 for GB model. We also implemented a two-layer fully connected neural network (NN) with a number of hidden units set to 100 for each layer.

We compared several existing EHR-based models. As in the original paper \cite{Miotto2016-hn}, DP was applied to reduce the dimensionality of EHR codes by training a denoising autoencoder with defaulted 500 hidden units followed by RF classifier from scikit-learn with 100 trees. For SiCNMF, we chose the number of factorization rank as 20 to avoid local minima which is suggested in the original paper. 

GRAM learns the ICD-9 embedding from a 5-level taxonomy called the Clinical Classification Software (CCS)    (\url{https://www.hcup-us.ahrq.gov/toolssoftware/ccs/AppendixCMultiDX.txt}). It then uses the ICD-9 embedding to project binary patient ICD-9 code vector onto a dense low-dimensional vector, which in turn serves as input features to a neural network for specific classification tasks. The original GRAM links the self-attention graph to a recurrent neural network (RNN) to do sequential diagnosis prediction. Since we performed only binary predictions on each target disease or outcome (i.e., CHD, COPD, and Diabetes-Insulin) by collapsing the time points, there is no recurrent unit. Therefore, the RNN (with one time point) reduced to a feedforward neural network. We use 200 attention weights and 100 embedding dimensions to represent each ICD code, and 100 one hidden layer with 200 hidden units in the feedforward network.


\subsection{Supplementary Figures}
\label{sec:complte_figures}

\begin{figure*}[t]
\centering\includegraphics[width=\textwidth]{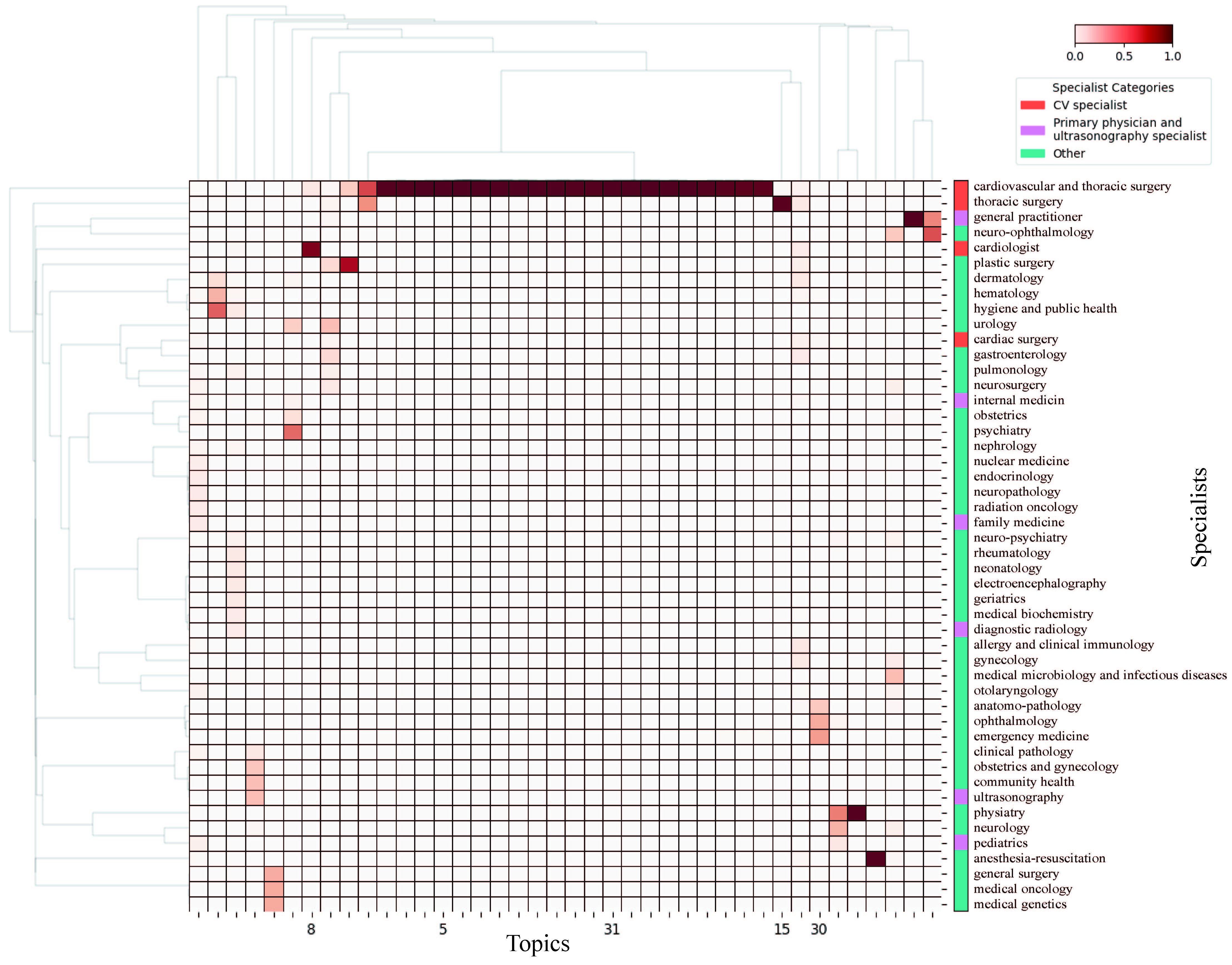}
\caption{The original specialist topic plot for CHD dataset with with complete 48 specialists. Side bar shows the corresponding category for each medical speciality. 
}
\label{fig:specialist_topic_CHD}
\end{figure*}

\begin{figure*}[!t]
\centering
\includegraphics[width=0.78\textwidth]{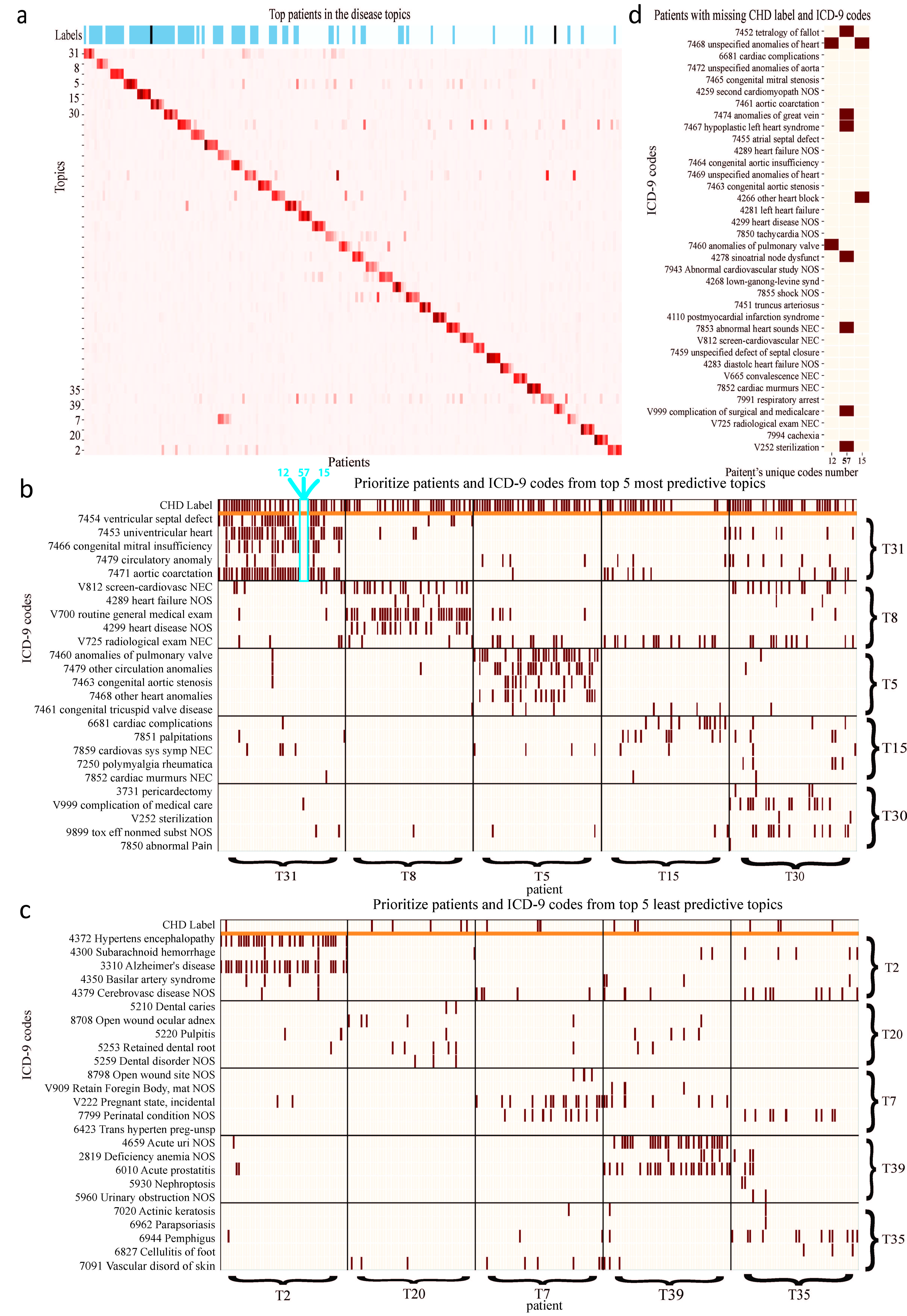}
\caption{\small{Prioritized patients by disease topic mixture from train data of CHD dataset. \textbf{a}. The top 5 patients under each topic. The patients with high proportions for the top positively and top negatively predictive topics are identified as high-risk patients and low-risk patients. 
\textbf{b}. Top ICD-9 codes of the high-risk patients. The top 5 ICD-9 codes (rows) from each topic were displayed for the top 50 patients (columns) under each topic. 
\textbf{c}. Top ICD-9 codes of the low-risk patients.}
\textbf{d}. The ICD-9 codes of the patients (12, 57, 15) highlighted in panel b.} \label{fig:patient_topic_plot}
\end{figure*}

\begin{figure*}[t]
\centering\includegraphics[width=\textwidth]{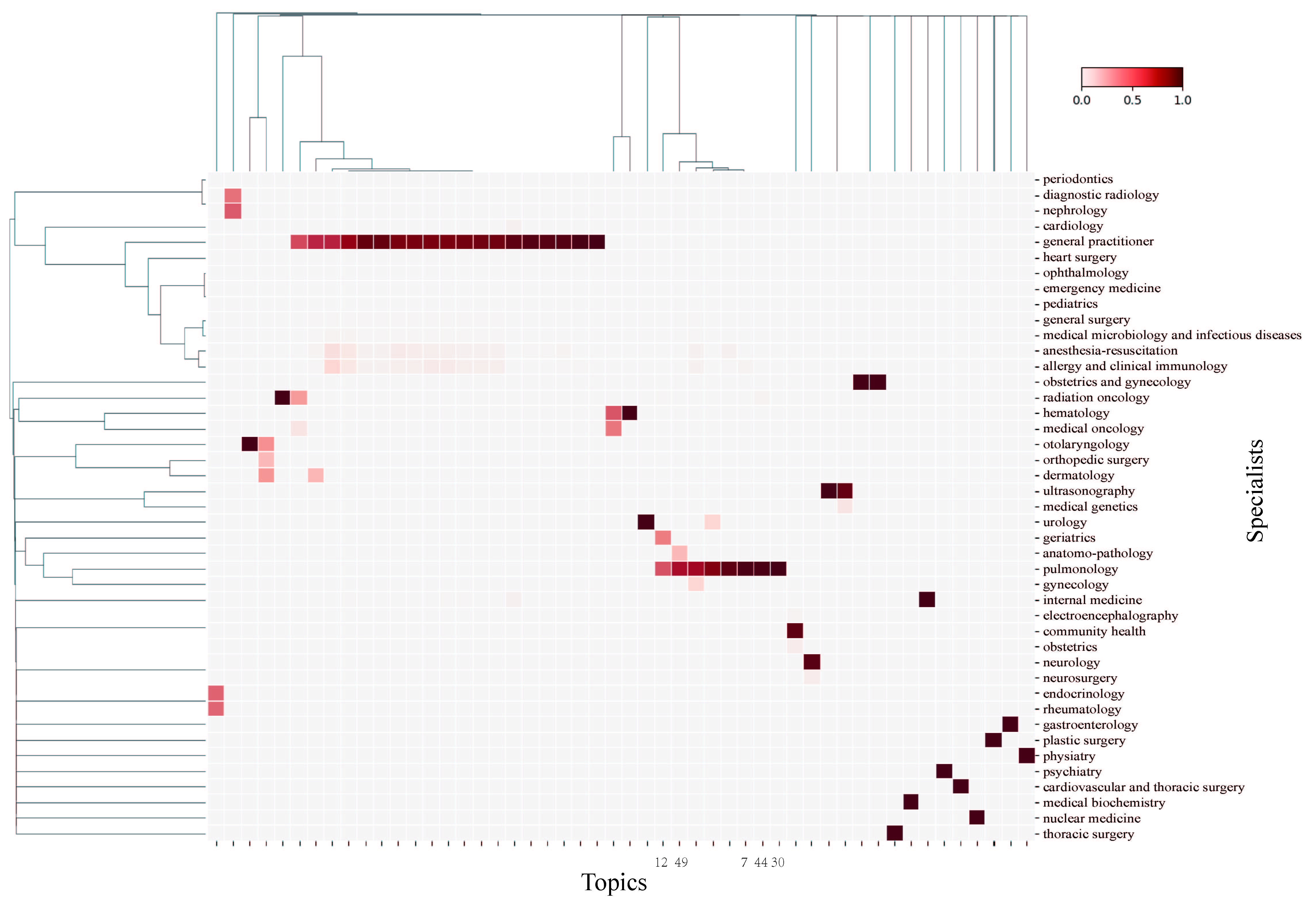}
\caption{The original specialist topic plot for COPD patients of PopHR database with complete 43 specialists.
}
\label{fig:specialist_topic_COPD}
\end{figure*}

\begin{figure*}[t]
\centering\includegraphics[width=\textwidth]{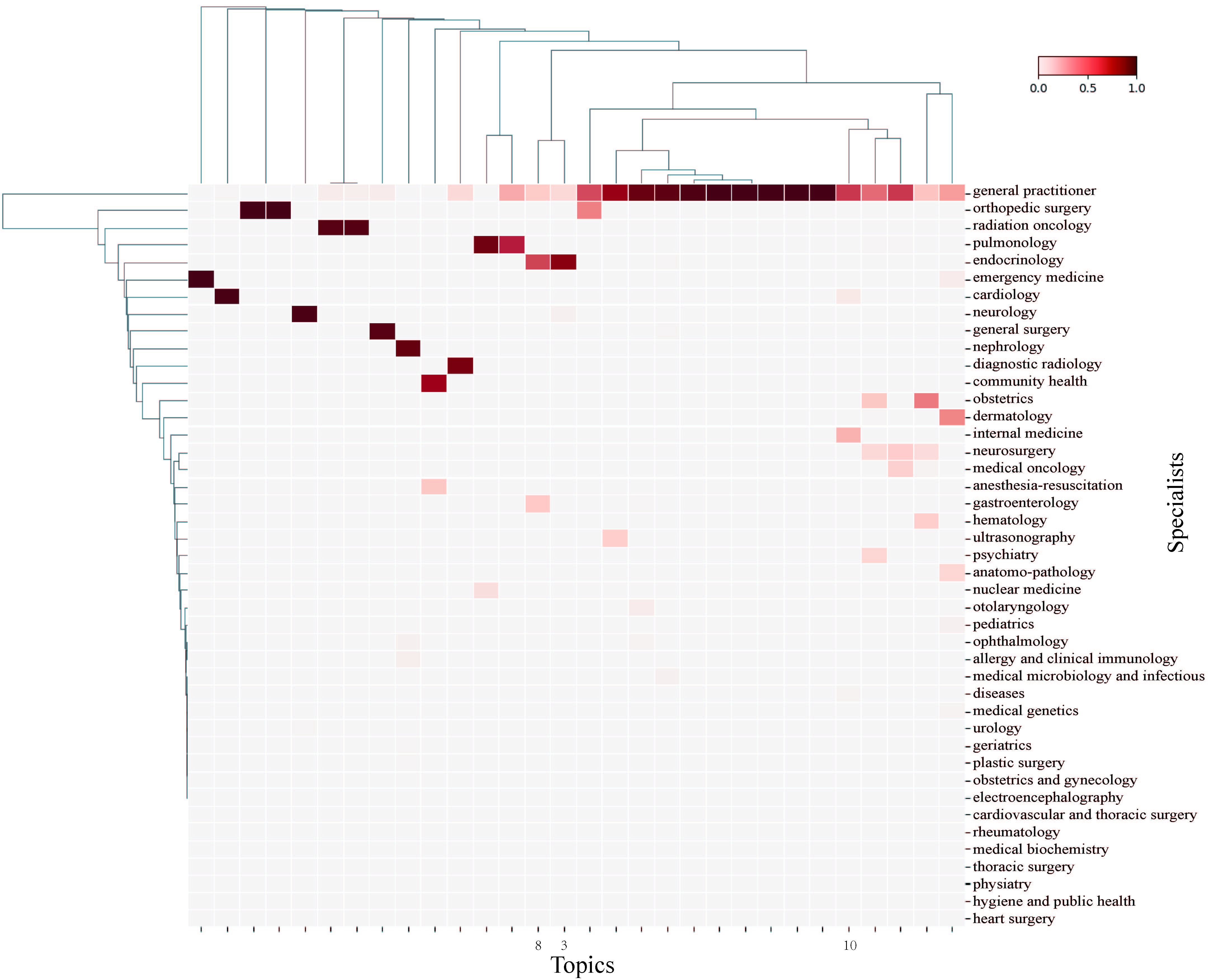}
\caption{The original specialist topic plot for diabetes patients of PopHR database with complete 43 specialists.
}
\label{fig:specialist_topic_diabete}
\end{figure*}

\clearpage
\pagebreak
\section{\mxr~Model Full derivation}
\label{sec:appendix:derivation}

\subsection{Derivation of \mxr~model likelihood} 
\label{sec:appendix_full_likelihood}

The full joint-likelihood for \mxr~  model(Fig.~\ref{fig:mixehr_model}) is:

\begin{align}
    p(\mf{z},\mf{b},\mf{x},\mf{y},\mf{g},\mf{w},\bm{\theta},\bm{\eta},\bm{\beta}) =& p(\bm\theta)p(\mf{z}\mid\bm\theta)p(\mf{b}\mid \mf{z},\bm\beta_k)p(\mf{x}\mid \mf{b},\mf{z},\bm\eta_{kt})p(\bm\beta_k)p(\bm\eta_{kt})\nonumber\\
    &p(\mf{y}\mid \mf{g})p(\mf{g}\mid \mf{z},\mf{w})p(\mf{w}) 
\end{align}
where the joint-likelihood involving the response latent variables ($\mf{y},\mf{g},\mf{w}$) from the supervised component of \mxr~ model  is:

\begin{align}
    p(\mf{y},\mf{g},\mf{w} \mid \mf{z},\tau) =& p(\mf{y} \mid \mf{g}) p(\mf{g} \mid \mf{w}, \mf{z}) p(\mf{w} \mid \tau) \nonumber\\
    =& p(\mf{w} \mid \tau)\prod_j^D  p(y_j \mid g_j) p(g_j \mid \mf{w}, \bar{\mf{z}}_j) \nonumber\\
    & (\frac{1}{2\pi})^{K/2}(\tau)^{1/2}\exp(-\frac{\tau \mf{w}^\intercal \mf{w}}{2})
\end{align}

For the unsupervised learning component of \mxr~model, we obtain marginal likelihood by integrating out $\bm\theta,\bm\eta,\bm\beta$ due to the conjugacy properties of the Dirichlet and multinomial distributions:

\begin{align}  
    p(\mf{z},\mf{b},\mf{x},\mf{g}) =& \int p(\bm{\theta})p(\mf{z}\mid\bm{\theta})d\bm{\theta} \times \int p(\mf{b}\mid \mf{z}, \bm{\beta_k} )p(\bm{\beta}_k)
    d\bm{\beta}_k \times 
    \int p(\mf{x}\mid \mf{b},\mf{z},\bm{\eta}_{kt})
    p(\bm{\eta}_{kt}) d\bm{\eta}_{kt}\nonumber\\
    =& p(\mf{z})p(\mf{b}\mid \mf{z})p(\mf{x}\mid \mf b, \mf z)
    \nonumber\\
    =& \prod_j^D \frac{\Gamma(\sum_k\alpha_k)}{\prod_k\Gamma(\alpha_k)}\frac{\prod_k\Gamma(\alpha_k+\sum_i^{M_j}[z_{ij}=k])}{\Gamma(\sum_k\alpha_k+\sum_i^{M_j}[z_{ij}=k])} \nonumber\\
    &\times \prod_k^K\frac{\Gamma(\sum_t\iota_t)}{\prod_t\Gamma(\iota_t)}\frac{\prod_t\Gamma(\iota_t+\sum_j^D\sum_i^{M_j}[z_{ij}=k,b_{ij}=t])}{\Gamma(\sum_t\iota_t+\sum_j^D\sum_i^{M_j}[z_{ij}=k,b_{ij}=t])}\nonumber\\
    &\times\prod_k^K\prod_t^T\frac{\Gamma(\sum_w\zeta_{kw})}{\prod_w\Gamma(\zeta_{kw})}\frac{\prod_w\Gamma(\zeta_{kw}+\sum_j^D\sum_i^{M_j}[b_{ij}=t,z_{ij}=k,x_{ij}=w])}{\Gamma(\sum_w\zeta_{kw}+\sum_j^D\sum_i^{M_j}[b_{ij}=t,z_{ij}=k,x_{ij}=w])}\nonumber\\
    =& \prod_j^D \frac{\Gamma(\sum_k\alpha_k)}{\prod_k\Gamma(\alpha_k)}\frac{\prod_k\Gamma(\alpha_k+n_{jk})}{\Gamma(\sum_k\alpha_k+n_{jk})} \nonumber\\
    &\times \prod_k^K\frac{\Gamma(\sum_t\iota_t)}{\prod_t\Gamma(\iota_t)}\frac{\prod_t\Gamma(\iota_t+m_{tk})}{\Gamma(\sum_t\iota_t+m_{tk})}\nonumber\\
    &\times\prod_k^K\prod_t^T\frac{\Gamma(\sum_w\zeta_{kw})}{\prod_w\Gamma(\zeta_{kw})}\frac{\prod_w\Gamma(\zeta_{kw}+p_{ktw})}{\Gamma(\sum_w\zeta_{kw}+p_{ktw})}  
    \label{eq:likelihood_unsupervised}
\end{align}
where ($n_{jk}, m_{tk},p_{twk}$) are the sufficient statistics.

\begin{align}
    n_{jk}&=\sum_i^{M_j}[z_{ij}=k]\nonumber\\
    m_{tk}&=\sum_j^D\sum_i^{M_j}[z_{ij}=k, b_{ij}=t]
    \nonumber\\
    p_{twk}&=\sum_j^D\sum_i^{M_j}[z_{ij}=k, b_{ij}=t,x_{ij}=w] 
\end{align}

We then can calculate the likelihood of unsupervised component of \mxr~model given the following analytical  expressions for each distribution of ($\mf z, \mf b, \mf x$):

\begin{align}
    p(\mf z) &= \prod_j^D \int \frac{\Gamma(\sum_k\alpha_k)}{\prod_k\Gamma(\alpha_k)}\prod_k^K\theta_{jk}^{\alpha_k-1}\prod_i^{M_j}\theta_{jk}^{[z_{ij}=k]}d\bm\theta_j\nonumber\\
    =&\prod_j^D \int\frac{\Gamma(\sum_k\alpha_k)}{\prod_k\Gamma(\alpha_k)}\prod_k^K\theta_{jk}^{\alpha_k+\sum_i^{M_j}[z_{ij}=k]-1}d\bm\theta_j\nonumber\\
    =&\prod_j^D \frac{\Gamma(\sum_k\alpha_k)}{\prod_k\Gamma(\alpha_k)}\frac{\prod_k\Gamma(\alpha_k+\sum_i^{M_j}[z_{ij}=k])}{\Gamma(\sum_k\alpha_k+\sum_i^{M_j}[z_{ij}=k])} 
\end{align}

\begin{align}
    p(\mf b\mid \mf z) &= \prod_k^K\int\frac{\Gamma(\sum_t\iota_t)}{\prod_t\Gamma(\iota_t)}\prod_t^T\beta_{kt}^{\iota_t-1}\prod_j^D\prod_i^{M_j}\beta_{kt}^{[z_{ij}=k,b_{ij}=t]}d\bm\beta_k\nonumber\\
    &= \int\prod_k^K\frac{\Gamma(\sum_t\iota_t)}{\prod_t\Gamma(\iota_t)}\prod_t^T\beta_{kt}^{\iota_t+\sum_j^D\sum_i^{M_j}[z_{ij}=k,b_{ij}=t]-1}d]\bm\beta_k\nonumber\\
    &= \prod_k^K\frac{\Gamma(\sum_t\iota_t)}{\prod_t\Gamma(\iota_t)}\frac{\prod_t\Gamma(\iota_t+\sum_j^D\sum_i^{M_j}[z_{ij}=k,b_{ij}=t])}{\Gamma(\sum_t\iota_t+\sum_j^D\sum_i^{M_j}[z_{ij}=k,b_{ij}=t])} 
\end{align}

\begin{align}
    p(\mf x\mid \mf b, \mf z) &= \prod_k^K\prod_t^T\int\frac{\Gamma(\sum_w\zeta_{kw})}{\prod_w\Gamma(\zeta_{kw})}\prod_w\eta_{ktw}^{\zeta_{kw}-1}\prod_j^D\prod_i^{M_j}\eta_{ktw}^{[b_{ij}=t,z_{ij}=k,x_{ij}=w]}d\bm\eta_{kt}\nonumber\\
    =&\prod_k^K\prod_t^T\int\frac{\Gamma(\sum_w\zeta_{kw})}{\prod_w\Gamma(\zeta_{kw})}\prod_w\eta_{ktw}^{\zeta_{kw}+\sum_j^D\sum_i^{M_j}[b_{ij}=t,z_{ij}=k,x_{ij}=w]-1}d\bm\eta_{kt}\nonumber\\
    =&\prod_k^K\prod_t^T\frac{\Gamma(\sum_w\zeta_{kw})}{\prod_w\Gamma(\zeta_{kw})}\frac{\prod_w\Gamma(\zeta_{kw}+\sum_j^D\sum_i^{M_j}[b_{ij}=t,z_{ij}=k,x_{ij}=w])}{\Gamma(\sum_w\zeta_{kw}+\sum_j^D\sum_i^{M_j}[b_{ij}=t,z_{ij}=k,x_{ij}=w])}
\end{align}

\subsection{Derivation of conditional distribution $z_{ij}=k$}
\label{sec:appendix_conditional}


In order to derive the conditional distribution $p(z_{ij} = k \mid \mf{z}_{(-ij)}, \mf b, \mf x, \mf g)$ which excludes specific $z_{ij}$ and ($\mf b, \mf x, \mf g$), we drop the  constant variables ($\alpha,  \iota, \zeta$), and also drop terms that depend on $(i,j)$ to get the conditional distribution:

\begin{align}  \label{eq:conditional dist gamma}
    p(z_{ij} = k & \mid \mf{z}_{(-ij)}, \mf b, \mf x, \mf g) =  \frac{p(z_{ij}, \mf{z}_{(-ij)},  \mf b,  \mf x,  \mf g)} {\sum_{k^{\prime}}^K p(z_{ij}, \mf{z}_{(-ij)}, \mf b, \mf x, \mf g)}\nonumber \\ 
    \propto & p(z_{ij}, \mf{z}_{(-ij)} \mid \alpha) p(\mf{b} \mid z_{ij}, \mf{z}_{(-ij)}, \iota) p(\mf{x} \mid z_{ij}, \mf{z}_{(-ij)}, \mf{b},\zeta_k) p(\mf{g}\mid z_{ij}, \mf{z}_{(-ij)}, \mf{w}) \nonumber \\
    \propto & (\alpha_{k} + n_{jk}^{-(i,j)})
    \frac{\iota_{b_{ij}} + m_{kb_{ij}}^{-(i,j)}}
    {\sum_t \iota_t + m_{kt}^{-(i,j)}} \frac{\zeta_{kx_{ij}} + p_{kb_{ij}x_{ij}}^{-(i,j)}} {\sum_w \zeta_{kw} + p_{kb_{ij}w}^{-(i,j)}}
    \nonumber \\
    & \times\exp\big\{\mf{w}^\intercal(\frac{1}{M_j}(\sum_{i^\prime \neq i} \mf{z}_{i^\prime j} + \mf{z}_{ij}))g_j -\frac{1}{2} \mf{w}^\intercal (\frac{1}{M_j^2} (\sum_{n\neq i}^{M_j} \sum_{m \neq n}^{M_j} \mf{z}_{nj} \mf{z}_{mj}^\intercal +
    \sum_{n\neq i}^{M_j} \mbox{diag}(\mf{z}_{nj})
    \nonumber \\
    & + \sum_{m\neq i}^{M_j} \mf{z}_{ij} \mf{z}_{mj}^\intercal +
    \sum_{n\neq i}^{M_j} \mf{z}_{nj} \mf{z}_{ij}^\intercal +
    \mbox{diag}(\mf{z}_{ij}))) \mf{w}\big\} \nonumber\\
    \propto & (\alpha_{k}+n_{jk}^{-(i,j)})\frac{\iota_{b_{ij}}+m_{kb_{ij}}^{-(i,j)}}{m_{k.}^{-(i,j)}+\sum_t\iota_t}\frac{\zeta_{kx_{ij}}+p_{kb_{ij}x_{ij}}^{-(i,j)}}{p_{kb_{ij}.}^{-(i,j)}+\sum_w\zeta_{kw}} \nonumber \\
    & \exp\big\{\frac{\mf{w}^\intercal \mf{z}_{ij} g_j}{M_j} -\frac{1}{2 M_j^2} \mf{w}^\intercal(\sum_{m\neq i}^{M_j} \mf{z}_{ij} \mf{z}_{mj}^\intercal +
    \sum_{n\neq i}^{M_j} \mf{z}_{nj} \mf{z}_{ij}^\intercal + \mbox{diag}(\mf{z}_{ij})) \mf{w}\big\}
\end{align}
where the notation $-(i,j)$ indicates that the term $(i,j)$ is excluded and $\mf{z}_{ij}$ is a \textit{one-hot} $K$-dimensional binary vector that sets the $k$th topic equal to 1 and remaining values equal to zero.


\subsection{Derivation of collapsed variational Bayesian inference (JCVB0)}
\label{sec:appendix_full_derivation}
To approximate the posterior distributions, we perform variational inference and obtain the ELBO $\mathcal{L}(\bm\Theta)$ in Eq. \eqref{eq:elbo} by using Jensen's inequality: 
\begin{align}
    \log p(\mf x,\mf b,\mf y) \geq & \mathbb{E}_{q(\mf z,\mf g,\mf w)} [\log\frac{p(\mf z,\mf g,\mf w,\mf b,\mf x,\mf y)}
    {q(\mf z,\mf g,\mf w)}] \nonumber\\
    =& \mathbb{E}_{q(\mf z,\mf g,\mf w)} [\log p(\mf z,\mf g,\mf w,\mf b, \mf x,\mf y)] - \mathbb{E}_{q(\mf z,\mf g,\mf w)} [\log q(\mf z,\mf g,\mf w)] = \mathcal{L}(\bm\Theta) 
\end{align}

The ELBO can be broken down using all variables in the \mxr~ model with their respective expectations:

\begin{align}
     \mathcal{L}(\bm\Theta) & = \mathbb{E}_{q(\mf z,\mf g,\mf w)} [\log p(\mf z,\mf g,\mf w,\mf b,\mf x,\mf y)] - \mathbb{E}_{q(\mf z,\mf g,\mf w)} [\log q(\mf z,\mf g,\mf w)] \nonumber \\
     & = \mathbb{E}_{q(\mf z)} [\log p(\mf z \mid \alpha)] +
     \mathbb{E}_{q(\mf z)} [\log p(\mf b \mid \mf z, \iota)] + \mathbb{E}_{q(\mf z)} [\log p(\mf x \mid \mf b, \mf z, \zeta)]\nonumber \\ 
     +& \mathbb{E}_{q(\mf z,\mf g,\mf w)} [\log p(\mf g \mid \mf z, \mf w)]
     +\mathbb{E}_{q(\mf w)} [\log p(\mf w \mid \tau)]
     +\mathbb{E}_{q(\mf g)} [\log p(\mf y \mid \mf g)] \nonumber \\
     -&\mathbb{E}_{q(\mf z)} [\log q(\mf z \mid \bm{\gamma})]
     -\mathbb{E}_{q(\mf g)} [\log q(\mf g \mid \bm{\lambda})]
     -\mathbb{E}_{q(\mf w)} [\log q(\mf w \mid \mf m, \mf S)] 
\end{align}
where the expectations of the above terms are: 
\begin{align}
    \mathbb{E}_{q(\mf z)}&[\log p(\mf z \mid \alpha)] =  \sum_j^D \mathbb{E}_{q(\mf z)}[ \log\Gamma(\sum_k \alpha_k) - \sum_k \log\Gamma(\alpha_k) \nonumber \\
     &+\sum_k \log \Gamma(\alpha_k+\sum_i^{M_j}[z_{ij}=k])
     - \log \Gamma (\sum_k\alpha_k+\sum_i^{M_j}[z_{ij}=k]) \nonumber \\
    = & \sum_j^D \log\Gamma(\sum_k \alpha_k) - \sum_k \log\Gamma(\alpha_k)  +\sum_k \log \Gamma(\alpha_k + E_q[n_{jk}]) - \log \Gamma (\sum_k\alpha_k + E_q[n_{jk}])
\end{align}

\begin{align}
    \mathbb{E}_{q(\mf z)}& [\log p(\mf b \mid \mf z, \iota)] = \sum_k^K \log \Gamma(\sum_t\iota_t) - \sum_t \log\Gamma(\iota_t) + \sum_t \log \Gamma(\iota_t + \sum_j^D\sum_i^{M_j}[b_{ij}=t] \gamma_{ijk}) \nonumber \\
    & - \log\Gamma(\sum_t\iota_t+\sum_j^D\sum_i^{M_j}[b_{ij}=t]\gamma_{ijk}) \nonumber \\
     = & \sum_k^K \log \Gamma(\sum_t\iota_t) - \sum_t \log\Gamma(\iota_t) + \sum_t \log \Gamma(\iota_t + E_q[m_{tk}]) \nonumber\\
     &- \log\Gamma(\sum_t\iota_t + E_q[m_{tk}])
\end{align}

\begin{align}
    \mathbb{E}_{q(\mf z)} [\log p(\mf x \mid \mf b, \mf z, \zeta)] =& \sum_k^K \sum_t^T \log\Gamma(\sum_w\zeta_{kw}) - \sum_w \log\Gamma(\zeta_{kw}) \nonumber\\
    &+ \sum_w \log\Gamma(\zeta_{kw} 
    + \sum_j^D\sum_i^{M_j}[b_{ij}=t,x_{ij}=w]\gamma_{ijk}) \nonumber \\
    & -\log\Gamma(\sum_w\zeta_{kw}+\sum_j^D\sum_i^{M_j}[b_{ij}=t,x_{ij}=w] \gamma_{ijk}) \nonumber \\
    =& \sum_k^K \sum_t^T \log\Gamma(\sum_w\zeta_{kw}) - \sum_w \log\Gamma(\zeta_{kw}) + \sum_w \log \Gamma(\zeta_{kw} + E_q[p_{twk}])  \nonumber \\
    & -\log\Gamma(\sum_w\zeta_{kw} + E_q[p_{twk}])
\end{align}

\begin{align}
    \mathbb{E}_{q(\mf z,\mf g,\mf w)} [\log p(\mf g \mid \mf z, \mf w)]  =& \sum_j \mathbb{E}_{q(\mf z,\mf g, \mf w)} [\log p(g_j \mid \mf w, \bar{\mf{z}}_j)] \nonumber\\
    =& \sum_j \mathbb{E}_{q(\mf z,\mf g, \mf w)} [\log \mathcal{N}(g_j \mid \mf{w}^\intercal \bar{\mf{z}}_j, 1)]\nonumber\\ 
     =& \sum_j^D -\frac{1}{2}\log 2\pi - \frac{1}{2}
    \mathbb{E}_{q(\mf z,\mf g, \mf w)}[(g_j-\mf{w}^\intercal \bar{\mf{z}}_j)^2]
    \nonumber\\ 
    =& \sum_j^D -\frac{1}{2}\log 2\pi - \frac{1}{2}
    \mathbb{E}_{q(\mf z,\mf g, \mf w)}[g_j^2 -2 \mf{w}^\intercal  \bar{\mf{z}}_j g_j + \mf{w}^\intercal \bar{\mf{z}}_j  \bar{\mf{z}}_j^\intercal \mf{w}] \nonumber\\ 
     =& \sum_j^D -\frac{1}{2}\log 2\pi - \frac{1}{2}
    \mathbb{E}_{q(\mf g)}[g_j^2]+\mathbb{E}_{q(\mf z,\mf g, \mf w)}[\mf{w}^\intercal \bar{\mf{z}}_j g_j] - \frac{1}{2} \mathbb{E}_{q(\mf z, \mf w)}[\mf{w}^\intercal \bar{\mf{z}}_j \bar{\mf{z}}_j^\intercal \mf{w}] \nonumber \\
    =& \sum_j^D -\frac{1}{2}\log 2\pi - \frac{1}{2}
    \mathbb{E}_{q(\mf g)}[g_j^2] + \mathbb{E}_{q(\mf w)}[\mf w]^\intercal \mathbb{E}_{q(\mf z)}[\bar{\mf{z}}_j] \mathbb{E}_{q(\mf g)}[g_j] \nonumber\\
    &- \frac{1}{2} \mathbb{E}_{q(\mf w)}\big[ \mf{w}^\intercal \mathbb{E}_{q(\mf z)}[\bar{\mf{z}}_j \bar{\mf{z}}_j^\intercal] \mf w\big]
\end{align}

\begin{align}
    \mathbb{E}_{q(\mf w)}[\log p(\mf w \mid \tau)] &= \mathbb{E}_{q(\mf w)}[\log \prod_k^K (\frac{1}{2\pi})^{K/2}(\tau_k)^{1/2}\exp(-\frac{\tau_k w_k^2}{2})] \nonumber \\
    & = \mathbb{E}_{q(\mf w)}[-\frac{K}{2} \log 2\pi + \frac{1}{2}\log\tau_k - \frac{\tau_k w_k^2}{2}] \nonumber \\
    & = -\frac{K}{2}\log 2\pi + \frac{1}{2}\log\tau_k - \frac{\tau_k \mathbb{E}_{q(\mf w)}[w_k^2]}{2} \nonumber \\
    & = -\frac{K}{2}\log 2\pi + \frac{1}{2}\log\tau_k - \frac{\tau_k(m_k^2 + S_{kk})}{2}
\end{align}

\begin{align}
    \mathbb{E}_{q(\mf g)}[\log p(\mf y \mid \mf g)] = \sum_j \mathbb{E}_{q(\mf g)} [\log p(y_j \mid g_j)] = 0 
\end{align}

\begin{align}
    \mathbb{E}_{q(\mf z)}[\log q(\mf z \mid \bm \gamma)] = \sum_{ijk} \gamma_{ijk} \log \gamma_{ijk}
\end{align}

\begin{align}
    \mathbb{E}_{q(\mf g)}[\log q(\mf g \mid \bm{\lambda})] =& \sum_j^D \mathbb{E}_{q(\mf g)}[\log q(g_j \mid \lambda_j)] \nonumber \\
    =& \sum_j^D \mathbb{E}_{q(\mf g)}[\log \big\{\mathcal{TN}^+(g_j;\lambda_j, 1)^{y_j}
    \mathcal{TN}^-(g_j;\lambda_j, 1)^{1-y_j}\big\}] \nonumber \\
    =& \sum_j^D \mathbb{E}_{q(\mf g)}[\log \big\{\mathcal{N}(g_j;\lambda_j, 1)
    (\frac{1}{1-\Phi_j})^{y_j}(\frac{1}{\Phi_j})^{1-y_j}\big\}] \nonumber \\
    =& \sum_j^D \mathbb{E}_{q(\mf g)} [\log\mathcal{N}(g_j;\lambda_j, 1)] - \big\{y_j \log(1-\Phi_j) + (1-y_j)\log\Phi_j\big\} \nonumber \\
    =& \sum_j^D -\frac{1}{2}\log 2\pi - \frac{1}{2} \mathbb{E}_{q(g)}[g_j^2]
    + \mathbb{E}_{q(g)}[g_j] \lambda_j - \frac{1}{2}\lambda_j^2 \nonumber\\
    &+ y_j \log(1-\Phi_j) + (1-y_j)\log\Phi_j
\end{align}

\begin{align}
    \mathbb{E}_{q(\mf w)}[\log q(\mf w \mid \mf m, \mf S)] &= \mathbb{E}_{q(\mf w)}[-\frac{K}{2} \log 2\pi - \sum_k^K(\frac{1}{2}\log S_{kk} + \frac{(w_k-m_k)^2}{2 S_{kk}})] \nonumber \\
    & = -\frac{K}{2}\log 2 \pi - \frac{K}{2} - \sum_k^K \frac{1}{2} \log S_{kk}
\end{align}

We use mean-field factorization for variational inference. For the latent topic variable $\mf{z}$, we posit a multinomial distribution with variational parameter $\bm{\gamma}$:
\begin{align}
    q(\mf{z} \mid \bm{\gamma}) = \prod_{ijk} \gamma_{ijk}^{[z_{ij} =k]} {,\;}
    \log q(\mf{z} \mid \bm{\gamma}) = \sum_{i,j,k} [z_{ij} = k] \log \gamma_{ijk}
\end{align}

We can maximize the ELBO with respect to the variational parameter $\gamma_{ijk}$ by calculating the expectation $\mathbb{E}_{q(\mf{z}^{-i,j})}[\ln q( \mf{z}_{ij} | \bm{\gamma})]$. The expect value of $\gamma_{ijk}$ is computed over variables ($\mf{z}, \mf{b}, \mf{x}, \mf{g}$):
\begin{align}\label{eq:log gamma}
    \log \gamma_{ijk}&\propto \mathbb{E}_{q(\mf{z}^{-(i,j)}, \mf{g}, \mf{w})}[\log p(\mf{z}_{ij}=k \mid \mf{z}_{(-ij)}, \mf{b}, \mf{x}, \mf{g})] \nonumber\\
    &= \mathbb{E}_{q(\mf{z}^{-(i,j)}, \mf{g}, \mf{w})}[\log p(\mf{z}, \mf{b}, \mf{x}, \mf{g})]
    \nonumber \\
    &= \mathbb{E}_{q(\mf{z}^{-(i,j)}, \mf{g}, \mf{w})}[\log 
    p(\mf{z} \mid \alpha) + \log p(\mf{b} \mid \mf{z}, \iota) \notag \\ 
    &+ \log p(\mf{x} \mid \mf{z}, \mf{b}, \zeta_k) +\log p(\mf{g} \mid \mf{w}, \mf{z})]
\end{align}

For the unsupervised component of \mxr~model, we update the first three terms whereas ignoring the last term $\log p(\mf{g} \mid \mf{w}, \mf{z})$ in Eq. \eqref{eq:log gamma}.  The variational update of $\gamma_{ijk}$ is obtained by normalizing itself and taking the expectation of the conditional distribution in Eq. \eqref{eq:conditional dist gamma} (see Appendix~\ref{sec:appendix_conditional}):
\begin{align} \label{eq:unsupervised update gamma}
    \gamma_{ijk} &= (\alpha_{k}+E_{q(\mf z^{-(i,j)})} [n_{jk}^{-(i,j)}])\notag\\
    &\frac{\iota_{b_{ij}} + E_{q(\mf z^{-(i,j)})}[m_{kb_{ij}}^{-(i,j)}]}
    {E_{q(\mf z^{-(i,j)})} [m_{k.}^{-(i,j)}] + \sum_t\iota_t}
    \frac{\zeta_{kx_{ij}}+E_{q(\mf z^{-(i,j)})} [p_{kb_{ij}x_{ij}}^{-(i,j)}]}{E_{q(\mf z^{-(i,j)})} [p_{kb_{ij}.}^{-(i,j)}]+\sum_w\zeta_{kw}}
\end{align}
where the above expected sufficient statistics are calculated by:
\begin{equation}
    E_{\mf{z}^{-(i,j)}} [n_{jk}^{-(i,j)}]=\sum_{i^{\prime}\neq i}^{M_{j}}\gamma_{i'jk}
\end{equation}

\begin{equation}
    E_{\mf{z}^{-(i,j)}} [m_{b_{ij}k}^{-(i,j)}]=
    \sum_{j^{\prime}\neq j}^D\sum_i^{M_{j^{\prime}}}
    [b_{i{j^{\prime}}}=b_{ij}]
    \gamma_{ij'k}
\end{equation}

\begin{equation}
    E_{\mf{z}^{-(i,j)}} [p_{b_{ij}x_{ij}k}^{-(i,j)}]=
    \sum_{j^{\prime}\neq j}^D\sum_i^{M_{j^{\prime}}}[
    b_{i{j^{\prime}}}=b_{ij}, x_{i{j^{\prime}}}=x_{ij} ]\gamma_{ij'k}
\end{equation}



For the latent liability variable $\mf{g}$, we propose a truncated Gaussian distribution as variational distribution with parameter $\bm{\lambda}$:
\begin{align}
    & q(g_j \mid \lambda_j) = \begin{cases}
    \mathcal{TN}_+(g_j;\lambda_j, 1), & \text{if} \quad y_j=1. \\
    \mathcal{TN}_-(g_j;\lambda_j, 1), & \text{if} \quad y_j=0.
    \end{cases}  \nonumber \\
    &\propto \begin{cases}
    \mathbbm{1}(g_j>0) \exp\big\{-\frac{1}{2}g_j^2+ \mathbb{E}_{q(\mf w)} [\mf{w}^\intercal] \mathbb{E}_{q(\mf z)}[\bar{\mf {z}}_j] g_j \big\}, & \text{if}\quad y_j=1. \\
    \mathbbm{1}(g_j \leq 0) \exp\big\{-\frac{1}{2}g_j^2+ \mathbb{E}_{q(\mf w)} [\mf{w}^\intercal] \mathbb{E}_{q(\mf z)}[\bar{\mf {z}}_j \bm ] g_j \big\} , & \text{if}\quad y_j=0.
    \end{cases} 
\end{align}

\begin{align}
    & \log q(g_j \mid \mf{z}, \mf{w}) = \mathbb{E}_{q(\mf{z}, \mf{w})}[\log p(g_j \mid \bar{\mf {z}}_j\bm , \mf{w})] + \log p(y_j \mid g_j) \nonumber \\
    & = y_j \log \mathbbm{1}(g_j>0) + (1-y_j) \mathbbm{1}(g_j \leq 0) - \frac{1}{2}g_j^2 + \mathbb{E}_{q(\mf{w})}[\mf{w}^\intercal] \mathbb{E}_{q(\mf{z})}[\bar{\mf {z}}_j\bm ] g_j 
\end{align}
where the update of the variational parameter $\lambda_j$ is:

\begin{align}
   \lambda_j = \mathbb{E}_{q(\mf{w})}[\mf{w}^\intercal] \mathbb{E}_{q(\mf{z})}[\bar{\mf{z}}_j]
    = \mf{m}^\intercal \bar{\bm{\gamma}}_{j}
\end{align}

For the linear coefficients $\mf{w}$, we propose a multivariate Gaussian distribution with mean parameter $\mf{m}$ and covariance parameter $\mf{S}$:
\begin{align}
    q(\mf{w} \mid \mf{m}, \mf{S}) = \mathcal{N}(\mf{w}\mid \mf{m}, \mf{S}) 
\end{align}   

\begin{align}   
    \log  q(\mf{w} \mid  \mf{m},  \mf{S}) & = \mathbb{E}_{q(\mf{z}, \mf{g})}
    [\log p(\mf{g} \mid \bar{\mf{z}}, \mf{w})] + \log p(\mf{w} \mid \tau)]
    \nonumber \\
    & = \mf{w}^\intercal \mathbb{E}_{q(\mf{z})}[\bar{\mf {z}}_j] \mathbb{E}_{q(\mf{g})}[\mf{g}] - \frac{1}{2} \mf{w}^\intercal(\tau \mf{I} + \mathbb{E}_{q(\mf{z})}[\bar{\mf{z}}_j^\intercal 
    \bar{\mf{z}}_j])\mf{w}
\end{align}

where the expectation terms are calculated in Appendix~\ref{appendix:expectation_calculation}. The full-derivation of $\mf{w}^\intercal \mathbb{E}_{q(\mf{z})} [\bar{\mf{z}}_j^\intercal \bar{\mf{z}}_j]) \mf{w}$ in Eq. \eqref{eq:wzzw}. We therefore obtain the following updates for $\mf{m}$ and $\mf{S}$:
\begin{align} \label{eq:update m s}
    \mf{m} = \mf{S} \mathbb{E}_{q(\mf{z})}[\bar{\mf{z}}]  \mathbb{E}_{q(\mf{g})}[\mf{g}],\quad 
    \mf{S} = (\tau \mf{I} + \mathbb{E}_{q(\mf{z})}[\bar{\mf{z}}^\intercal \bar{\mf{z}}])^{-1}
\end{align}
where the expectation $ \mathbb{E}_{q(\mf{z})} [\bar{\mf{z}}_j^\intercal \bar{\mf{z}}_j])$ is derived in Eq.~\eqref{eq:zz}.

After we assign variational distributions for $\mf{g}$ and $\mf{w}$, we can update $\gamma_{ijk}$ using the supervised component of the MixEHR-S. Here, we take the expectation of the predictive likelihood $\log p(\mf{g} \mid \mf{w}, \mf{z}^{-(i,j)}, z_{ijk}=1,z_{ijk'}=0\forall{k'\ne k})]$ in Eq.~\eqref{eq:log gamma}:
\begin{align*}\label{eq:supervised update gamma}
    &\mathbb{E}_{q(\mf{z}^{-(i,j)})} [\log p(\mf{g} \mid \mf{w}, \mf{z}^{-(i,j)}, z_{ijk}=1,z_{ijk'}=0\forall{k'\ne k})]\\
    =& \mathbb{E}_{q(\mf{z}^{-(i,j)}, \mf{g}, \mf{w})}[\log \big( \frac{1}{\sqrt{2\pi}} \exp\big\{-\frac{(g_j - \mf{w}^\intercal \bar{\mf{z}}_j )^2}{2}\big\}] \big)\\
    \propto&  \mathbb{E}_{q(\mf{z}^{-(i,j)}, \mf{g}, \mf{w})}
    [\mf{w}^\intercal (\frac{1}{M_j}(\sum_{i^\prime \neq i} \mf{z}_{i^\prime j} + \mf{z}_{ij}))g_j -\frac{1}{2} \mf{w}^\intercal (\frac{1}{M_j^2}(
    \sum_{m\neq i}^{M_j} \mf{z}_{ij} \mf{z}_{mj}^\intercal +
    \sum_{n\neq i}^{M_j} \mf{z}_{nj} \mf{z}_{ij}^\intercal +
    \mbox{diag}(\mf{z}_{ij})) \mf{w}]\\
    \propto & \frac{\mf{w}^\intercal \mf{z}_{ij} \mathbb{E}_{q(\mf{g})}[g_j]}{M_j}-\frac{1}{2 M_j^2} \mathbb{E}_{q(\mf{z}^{-(i,j)},\mf{w})}\left[\mf{w}^\intercal(\mf{z}_{ij}\underbrace{\sum_{m\neq i}^{M_j}\mf{z}_{mj}^\intercal}_{\mf{z}^\top_{j/i}} +
    \underbrace{\sum_{n\neq i}^{M_j} \mf{z}_{nj}}_{\mf{z}_{j/i}} \mf{z}_{ij}^\intercal +
    \mbox{diag}(\mf{z}_{ij}))\mf{w}\right]\\
    =& \frac{m_k \mathbb{E}_{q(g_j)}[g_j]}{M_j} - \frac{1}{2 M_j^2} \mathbb{E}_{q(\mf{z}^{-(i,j)},\mf{w})}\left[\mf{w}^\intercal(
    \mf{z}_{ij}\mf{z}_{j/i}^{\intercal}+\mf{z}_{j/i}\mf{z}_{ij}^\intercal +
    \mbox{diag}(\mf{z}_{ij}))\mf{w}\right]\\
    =& \frac{m_k \mathbb{E}_{q(g_j)}[g_j]}{M_j} -
    \frac{1}{2 M_j^2} \mathbb{E}_{q(\mf{z}^{-(i,j)},\mf{w})}\left[w_k\mf{z}_{j/i}^\intercal\mf{w} + \mf{w}^\intercal\mf{z}_{j/i}w_k + w^2_k\right]\\
    =& \frac{m_k \mathbb{E}_{q(g_j)}[g_j]}{M_j} -
    \frac{1}{2 M_j^2} \mathbb{E}_{q(\mf{z}^{-(i,j)},\mf{w})}\left[
    2\sum_{m\ne i}\sum_{k'} w_kw_{k'}z_{mjk} + w^2_k\right]\\
    =& \frac{m_k \mathbb{E}_{q(\mf{g})}[g_j]}{M_j} -
    \frac{1}{2 M_j^2} 
    \left[2\sum_{m\ne i}\sum_{k'} (m_km_{k'} + S_{kk'})\gamma_{mjk} + m^2_k + S_{kk}\right]\\
    =& \frac{m_k \mathbb{E}_{q(g_j)}[g_j]}{M_j} -
    \frac{1}{2 M_j^2} 
    \left[2(m_k\mf{m}^\intercal\bm{\gamma}_{j/i} + \mf{S}_{k}\bm{\gamma}_{j/i})+m^2_k+S_{kk}\right]
\end{align*}
where $\bm{\gamma}_{j/i}$ indicates the sum of all terms except for the $i$th ICD-9 code, i.e. $\bm{\gamma}_{j/i} = \sum_{m\neq i}^{M_j} \bm{\gamma}_{ij}$ and $\mf{S}_k$ indicates the $k$th row of the covariance matrix $\mf{S}$. The expected value of $\mf{w}^\intercal\mf{w}$ is computed in Eq. \eqref{eq:exp_w}.

We thus add the supervised component to $\bm \gamma$ (see Eq.~\eqref{eq:unsupervised update gamma} and Eq. \eqref{eq:supervised update gamma}), obtaining full variational update for $\gamma_{ijk}$:
\begin{align}
    \gamma_{ijk} \propto & (\alpha_{k}+E_{q(z^{-(i,j)})} [n_{jk}^{-(i,j)}])\frac{\iota_{b_{ij}} + E_{q(z^{-(i,j)})}[m_{kb_{ij}}^{-(i,j)}]}
    {E_{q(z^{-(i,j)})} [m_{k.}^{-(i,j)}] + \sum_t\iota_t}
    \frac{\zeta_{kx_{ij}}+E_{q(z^{-(i,j)})} [p_{kb_{ij}x_{ij}}^{-(i,j)}]}{E_{q(z^{-(i,j)})} [p_{kb_{ij}.}^{-(i,j)}]+\sum_w\zeta_{kw}} \nonumber \\
    &\exp \left\{\frac{m_k \mathbb{E}_{q(g_j)}[g_j]}{M_j} -
    \frac{1}{2 M_j^2} 
    \left[2\left(m_k\mf{m}^\intercal\bm{\gamma}_{j/i} + \mf{S}_{k}\bm{\gamma}_{j/i}\right)+m^2_k+S_{kk}\right]\right\}
\end{align}

\subsection{Hyperparameters update} \label{sec:appendinx_hyperparameters}
We update the hyperparameters ($\alpha_k^{\ast}$, $\iota_t^{\ast}$, $\zeta_{wk}^{\ast}$) by maximizing the marginal likelihood under the variational expectations via empirical Bayes fixed point iteration method \cite{asuncion2009smoothing,minka2000estimating}:
\begin{align}\label{eq:topic_hyper_alpha}
    \alpha_k^{\ast} &= \frac{c_{\alpha}-1+\alpha_k\sum_j \Psi(\alpha_k + n_{jk}) - \Psi(\alpha_k)}
    {d_{\alpha}+\sum_j\Psi(\sum_k\alpha_k+n_{jk}) - \Psi(\sum_k\alpha_k)}
\end{align}

\begin{align}
\label{eq:topic_hyper_iota}
    \iota_t^{\ast} &= \frac{c_{\iota}-1+\iota_t\sum_k \Psi(\iota_t + m_{tk}) - \Psi(\iota_t)}
    {d_{\iota}+\sum_k\Psi(\sum_t\iota_t+m_{tk}) - \Psi(\sum_t\iota_t)}
\end{align}                                                                                
\begin{align}
\label{eq:topic_hyper_zeta}
    \zeta_{wk}^{\ast} &= \frac{c_{\zeta}-1+\zeta_{wk}\sum_t \Psi(\zeta_{wk} + p_{twk}) - \Psi(\zeta_{wk})}
    {d_{\zeta}+\sum_t\Psi(\sum_w\zeta_{wk}+p_{twk}) - \Psi(\sum_w\zeta_{wk})}
\end{align}
where $(c_{\alpha}, c_{\iota}, c_{\zeta}, d_{\alpha},  d_{\iota} d_{\zeta})$ are constant values. For the experiment, we chose the following initial value setting $(1, 0.001, 2, 10, 0.01, 100)$.

\subsection{Derivation of expectations of variational distributions}\label{appendix:expectation_calculation}
The expected value associated with the average of the topic assignments $\bar{\mf{z}}_j$ is:
\begin{equation}
    \mathbb{E}_{q(\mf z)}[\bar{\mf{z}}_j] = \bar{\bm{\gamma}}_j = \frac{1}{M_j} \sum_i^{M_j}\bm{\gamma}_{ij}
\end{equation}

The expectation of $\bar{\mf{z}}_j \bar{\mf{z}}_j^\intercal$ is:
\begin{align}\label{eq:zz}
    \mathbb{E}_{q(\mf{z})} [\bar{\mf{z}}_j \bar{\mf{z}}_j^\intercal] &= 
    \mathbb{E}_{q(\mf z)} \left[
    \begin{bmatrix}
       \frac{1}{M_j} \sum_i^{M_j} z_{ij1}\\
       \vdots \\
       \frac{1}{M_j} \sum_i^{M_j} z_{ijK}
    \end{bmatrix}
    \begin{bmatrix} \frac{1}{M_j} \sum_i^{M_j} z_{ij1} \dots  \frac{1}{M_j} \sum_i^{M_j} z_{ijK} \end{bmatrix}\right]
    \nonumber \\
    & = \mathbb{E}_{q(\mf z)}\left[ \begin{bmatrix}
    \frac{1}{M_j^2}\sum_i^{M_j} z_{ij1}^2 \dots  & \frac{1}{M_j^2}\sum_i^{M_j} z_{ij1} z_{ijK}  \\
    \vdots \ddots & \vdots \\
    \frac{1}{M_j^2}\sum_i^{M_j} z_{ijK} z_{ij1} \dots  & \frac{1}{M_j^2}\sum_i^{M_j} z_{ijK}^2
    \end{bmatrix}\right]
    \nonumber\\
    &= \frac{1}{M_j^2}\big(\sum_i^{M_j} \sum_{i^\prime}^{M_j} \mathbb{E}_{q(\mf z)}[\mf{z}_{ij}] \mathbb{E}_{q( \mf z)}[\mf{z}_{i^\prime j}^\intercal] +\sum_i^{M_j} \mbox{diag}(\mathbb{E}_{q(\mf{z})}[\mf{z}_{ij}])\big) \nonumber\\
    & = \frac{1}{M_j^2}\big(\sum_i^{M_j} \sum_{i^\prime}^{M_j} \bm{\gamma}_{ij} \bm{\gamma}_{i^\prime j}^\intercal +\sum_i^{M_j} \mbox{diag}(\bm{\gamma}_{ij})\big)
\end{align}

The expected value of $\mf{w}^\intercal \mathbb{E}_{q(\mf{z})}[\bar{\mf{z}}_j \bar{\mf{z}}_j^\intercal]\mf{w}$ given the variational distribution $q(\mf w)$ is:
\begin{align}  \label{eq:wzzw}
    \mathbb{E}_{q(\mf w)}\left[\mf{w}^\intercal \mathbb{E}_{q(\mf z)}[\bar{\mf{z}}_j \bar{\mf{z}}_j^\intercal]\mf{w}\right]&=
    \mathbb{E}_{q(\mf w)}\left[
    \begin{bmatrix} w_1 \dots w_K \end{bmatrix}
    \begin{bmatrix}
    \frac{1}{M_j^2}\sum_i^{M_j} \gamma_{ij1}^2 \dots  & \frac{1}{M_j^2}\sum_i^{M_j} \gamma_{ij1} \gamma_{ijK}  \nonumber \\
    \vdots \ddots & \vdots \\
    \frac{1}{M_j^2}\sum_i^{M_j} \gamma_{ijK}\gamma_{ij1} \dots  & \frac{1}{M_j^2}\sum_i^{M_j} \gamma_{ijK}^2
    \end{bmatrix}
    \begin{bmatrix}
       w_1\\
       \vdots \\
       w_K
    \end{bmatrix}
    \right]
    \nonumber \\
    &=
    \mathbb{E}_{q(\mf w)}\left[
    \begin{bmatrix} w_1 \dots w_K \end{bmatrix}
    \begin{bmatrix}
    \bar\gamma_{j1}^2 \dots  & \bar\gamma_{j1} \bar\gamma_{jK}  \nonumber \\
    \vdots \ddots & \vdots \\
    \bar\gamma_{jK}\bar\gamma_{j1} \dots  & \bar\gamma_{jK}^2  
    \end{bmatrix}
    \begin{bmatrix}
       w_1\\
       \vdots \\
       w_K
    \end{bmatrix}
    \right]
    \nonumber \\
    &=\mathbb{E}_{q(\mf w)}\left[\begin{bmatrix} \mf{w}^\intercal \bar{\bm{\gamma}}_{j.} \bar\gamma_{j1} 
    \dots \mf{w}^\intercal \bar{\bm{\gamma}}_{j.} \bar\gamma_{jK} \end{bmatrix}
    \begin{bmatrix}
       w_1\\
       \vdots \\
       w_K
    \end{bmatrix}
    \right]
    \nonumber \\
    &= \mathbb{E}_{q(\mf w)}\big[\sum_{k^\prime} w_{k^\prime} \mf{w}^\intercal \bar{\bm{\gamma}}_{j.} \bar\gamma_{jk^\prime}] \nonumber \\ 
    &=\sum_{k^\prime} \mathbb{E}_{q(\mf w)}\big[w_{k^\prime} \mf{w}^\intercal\big] \bar{\bm{\gamma}}_{j.} \bar\gamma_{jk^\prime} \nonumber \\
    &=\sum_{k^\prime}
    \begin{bmatrix}
       \mathbb{E}_{q(\mf w)}\big[w_{k^\prime} w_1\big] \\
       \vdots \\
       \mathbb{E}_{q(\mf w)}\big[w_{k^\prime} w_K\big]
    \end{bmatrix}^\intercal
    \bar{\bm{\gamma}}_{j.} \bar\gamma_{jk^\prime} \nonumber \\
    &=\sum_{k^\prime}
    \begin{bmatrix}
       m_{k^\prime} m_1 + S_{k^\prime 1}\\
       \vdots \\
       m_{k^\prime} m_K + S_{k^\prime K}
    \end{bmatrix}^\intercal
    \bar{\bm{\gamma}}_{j.} \bar\gamma_{jk^\prime} 
\end{align}
where the $\bar{\bm{\gamma}}_{j.} = \frac{1}{M_j} \sum_i \bm{\gamma}_j$ is a $K$-dimensional vector, each scalar value $\bar{\gamma}_{jk}$ represents the average value of $\gamma$ over ICD-9 codes for the $k$th topic and the $j$th patient. 

The expected value of latent liability variable $\mf{g}$ for each patient $j$ is given:
\begin{align}
    & \mathbb{E}_{q(\mf{g})}[g_j] = \begin{cases}
    \lambda_j + \phi_j / (1-\Phi_j), & \text{if}\quad y_j=1.\\
    \lambda_j - \phi_j / \Phi_j, & \text{if}\quad y_j=0.
    \end{cases} 
\end{align}
where $\phi_j = \phi(-\lambda_j)$ is the normal density and $\Phi_j = \Phi(-\lambda_j)$ is the cumulative distribution function (CDF) of the standard normal distribution. The corresponding expected value for $\mf w$ is:
\begin{align} \label{eq:exp_w}
    \mathbb{E}_{q(\mf w)}[\mf{w}] = \mf{m}, \quad
    \mathbb{E}_{q(\mf w)}[\mf{w}^\intercal \mf{w}] = \mathbb{E}_{q(\mf w)}[\sum_{k=1}^K w_k w_k] = 
    \sum_{k=1}^K m_k^2 + S_{kk} 
\end{align}

\subsection{Derivation of estimates for variational expectations of mixing proportions}

The approximations for mixing proportions ($\bm\theta$, $\bm\beta$, $\bm\eta$) can be calculated as follows \cite{teh2007collapsed}:

\begin{align}
    \hat{\bm{\theta}} &= \frac{\alpha_k + \mathbb{E}_{q(\mf z)}[n_{j.k}]}{\sum_{k^{\prime}} \alpha_{k^{\prime}} + \mathbb{E}_{q( \mf z)}[n_{j.{k^{\prime}}}]}\\ 
    \hat{\bm{\beta}} &= \frac{\iota_t + \mathbb{E}_{q( \mf z)}[m_{.kt}]}{\sum_{t^{\prime}} \iota_{t^{\prime}} + \mathbb{E}_{q( \mf z)}[m_{.kt^{\prime}}]}\\ 
    \hat{\bm{\eta}} &= \frac{\zeta_w + \mathbb{E}_{q(\mf z)}[p_{.ktw}]}{\sum_{w^{\prime}}  \zeta_{w^{\prime}} + \mathbb{E}_{q( \mf z)}[p_{.ktw^{\prime}}]} 
\end{align}
where the expected values of sufficient statistics $(n_{j.k}, m_{.kt}, p_{.ktw})$ are:
\begin{align} 
    \mathbb{E}_{q(\mf z)}[ n_{j.k}] &=\sum_i^{M_{j}} [z_{ij}=k] \\
    \mathbb{E}_q(\mf z)[m_{.kt}] &= \sum_{j^{\prime}}^D\sum_i^{M_{j^{\prime}}} [z_{ij^{\prime}}=k, b_{ij^{\prime}}=t]  \\
    \mathbb{E}_q(\mf z)[p_{.ktw}] &= \sum_{j^{\prime}}^D\sum_i^{M_{j^{\prime}}} [z_{ij^{\prime}}=k, b_{ij^{\prime}}=t, x_{ij^{\prime}}=w] 
\end{align}

\subsection{Derivation of predictive distribution}
Here, we show that the predictive distribution is a Bernoulli distribution when using a Gaussian response since the natural parameter $\mf{w}^\intercal \bar{\mf{z}}$ is identical to the mean parameter where $\bar{\mf{z}} \star$ and $\mf{y} \star$ represent the new data points and the predicted label respectively. The full derivation of the predictive distribution is:
\begin{align}
    p(\mf{y}\star \mid \bar{\mf{z}}\star, \mf{y}, \bar{\mf{z}}) & = \int_{-\infty}^{\infty} \int p(\mf{y}\star, \mf g, \mf w \mid \bar{\mf{z}}\star, \mf{y}, \bar{\mf{z}}) d\mf{w} d\mf{g} \nonumber \\
    & = \int_{-\infty}^{\infty} \int p(\mf{y}\star \mid \mf g) 
    p(\mf g \mid \mf w, \bar{\mf{z}}\star) p(\mf w \mid \mf y, \bar{\mf{z}}) d\mf{w} d\mf{g} \nonumber \\
    & \approx \int_{-\infty}^{\infty} \int p(\mf{y}\star \mid \mf g) p(\mf g \mid \mf w, \bar{\mf{z}}\star) q(\mf w) d\mf{w} d\mf{g} \nonumber \\
    & = \int_{-\infty}^{\infty} \int \mathbb{1}(\mf{g} > 0)^{\mf{y}\star} \mathbb{1}_((\mf{g} \leq 0)^{1 - \mf{y}\star} 
    \mathcal{N}(\mf{g} \mid \mf{w}^\intercal \bar{\mf{z}} \star, 1) 
    \mathcal{N}(\mf w \mid \mf m, \mf S) d\mf{w} d\mf{g} \nonumber \\
    & = \begin{cases} 
      \int_{0}^{\infty}  \mathcal{N}(\mf{g} \mid \mf{m}^\intercal \bar{\mf{z}} \star, 1 + \bar{\mf{z}} \star^\intercal \mf S \bar{\mf{z}} \star) d\mf{g}
      & \text{if}\quad  \mf{y} \star = 1 \\
      \int_{-\infty}^{0} \mathcal{N}(\mf{g} \mid \mf{m}^\intercal \bar{\mf{z}} \star, 1 + \bar{\mf{z}} \star^\intercal \mf S \bar{\mf{z}} \star) d\mf{g} 
      & \text{if}\quad  \mf{y} \star = 0   
    \end{cases} \nonumber \\
    & = \begin{cases} 
      1 - \Phi\bigg(\frac{-\mf{m}^\intercal \bar{\mf{z}} \star}{(1 + \bar{\mf{z}} \star^\intercal  \mf S \bar{\mf{z}} \star)^{\frac{1}{2}}}\bigg) & \text{if}\quad \mf{y} \star = 1 \\
      \Phi\bigg(\frac{-\mf{m}^\intercal \bar{\mf{z}} \star}{(1 + \bar{\mf{z}} \star^\intercal  \mf S  \bar{\mf{z}} \star)^{\frac{1}{2}}}\bigg) & \text{if}\quad \mf{y} \star = 0
    \end{cases} \nonumber \\
    & = \Phi\bigg(\frac{\mf{m}^\intercal \bar{\mf{z}} \star}{(1 + 
    \bar{\mf{z}} \star^\intercal  \mf S \bar{\mf{z}} \star
     )^{\frac{1}{2}}}\bigg)^{\mf{y} \star}
    \bigg(1 - \Phi\bigg(\frac{\mf{m}^\intercal \bar{\mf{z}} \star}{(1 + \bar{\mf{z}} \star^\intercal  \mf S \bar{\mf{z}} \star
    )^{\frac{1}{2}}}\bigg)\bigg)^{1 - \mf{y} \star}
    \nonumber \\
    & = \text{Bernoulli}\Bigg(\mf{y}\star \mid \Phi\bigg(\frac{\mf{m}^\intercal \bar{\mf{z}} \star}
    {(1 + \bar{\mf{z}} \star^\intercal  \mf S \bar{\mf{z}} \star)^{\frac{1}{2}}}\bigg)\Bigg)
\end{align}

\end{document}